\documentclass[10pt,twocolumn,letterpaper]{article}

\usepackage[pagenumbers]{iccv} % To force page numbers, e.g. for an arXiv version

\usepackage{soul}
\usepackage{bm}
\usepackage{nicefrac}
\usepackage{multirow}
\usepackage{booktabs}
\usepackage{arydshln}
\usepackage[normalem]{ulem}
\useunder{\uline}{\ul}{}

\definecolor{iccvblue}{rgb}{0.21,0.49,0.74}
\usepackage[pagebackref,breaklinks,colorlinks,allcolors=iccvblue]
{hyperref}
\usepackage{amssymb}
\usepackage{bbding}
\usepackage{float}
\usepackage{pifont}
\usepackage[accsupp]{axessibility}

\def\paperID{3739} % *** Enter the Paper ID here
\def\confName{ICCV}
\def\confYear{2025}

\title{Effective Training Data Synthesis for Improving MLLM Chart Understanding}

\author{
    Yuwei Yang$^{1}$, Zeyu Zhang$^{1}$, Yunzhong Hou$^{1}$, Zhuowan Li$^{4}$, Gaowen Liu$^{3}$, Ali Payani$^{3}$,\\
    Yuan-Sen Ting$^{2}$, Liang Zheng$^{1}$\\
    $^{1}$Australian National University \quad $^{2}$Ohio State University\\
    $^{3}$Cisco \quad $^{4}$Johns Hopkins University \\
    {\tt\small \{yuwei.yang, zeyu.zhang, yunzhong.hou, 
 liang.zheng\}@anu.edu.au} \quad {\tt\small ting.74@osu.edu} \\
 \quad {\tt\small \{gaowen.liu, ali.payani\}@cisco.com} \quad {\tt\small zli110@jhu.edu}
}

\begin{document}

\twocolumn[{% 
\renewcommand\twocolumn[1][]{#1}%
\maketitle  
\thispagestyle{plain} 
\vskip -8mm
\begin{figure}[H]                                    
\hsize=\textwidth
\centering
  % \fbox{\rule{0pt}{1.5in} \rule{0.9\linewidth}{0pt}}
   \includegraphics[width=\textwidth]{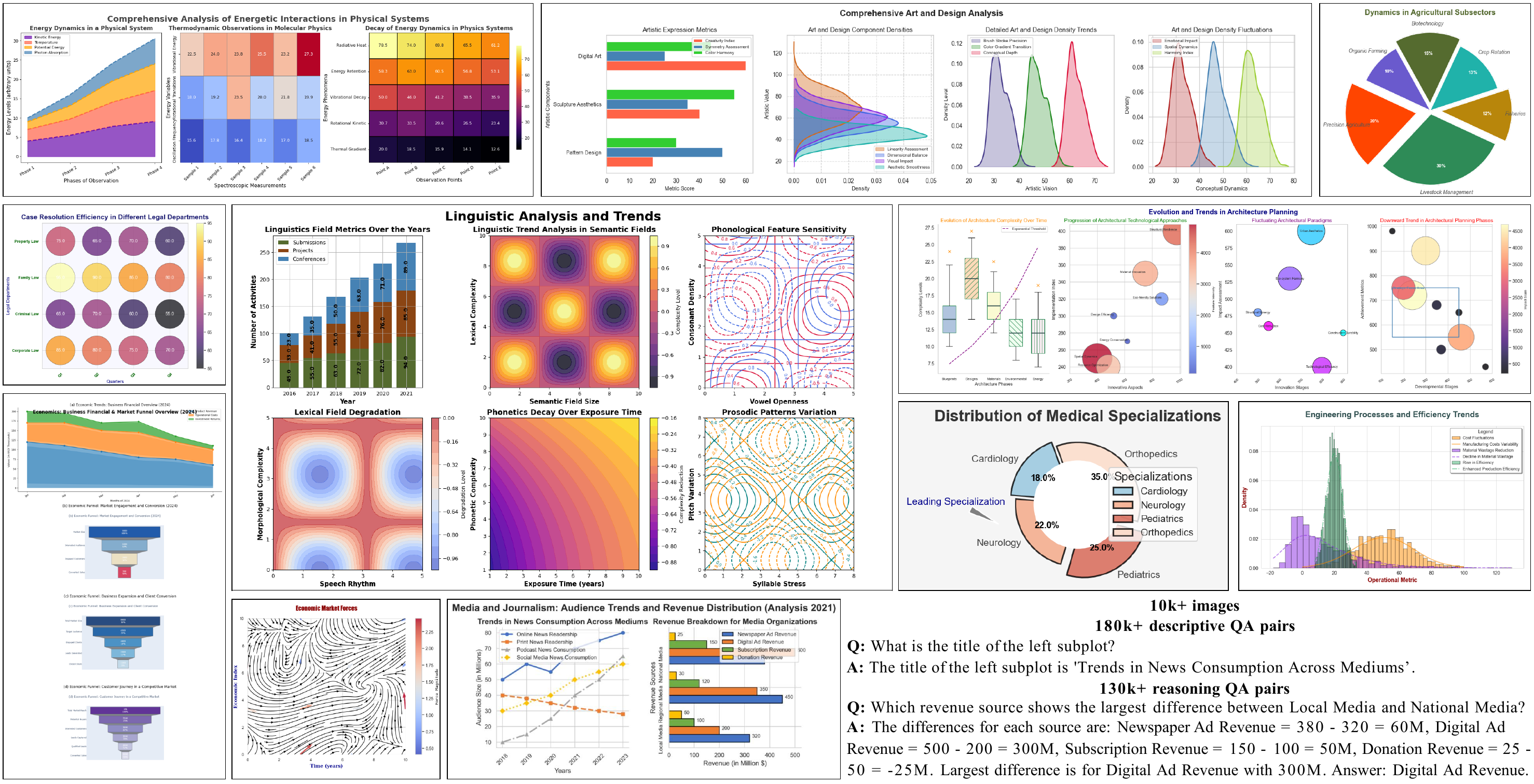}
   \vskip -1mm
   \caption{Overview of the effective chart dataset (ECD). The QA-referenced chart has been adjusted for readability. The ECD is composed of chart images and question-answer (QA) pairs and is intended as a training set for chart understanding. It includes 29 chart types and over 250 subplot combinations, offering more than 10k images and 300k QA pairs with high quality.} 
   \label{fig:motivation}
\end{figure}
}]
\setcounter{page}{1} 

\begin{abstract}
\renewcommand{\thefootnote}{}\footnote{*This work was supported in part by the Australian Research Council under Discovery Project DP210102801 and Future Fellowship FT240100820, as well as by the National Science Foundation under Grant No. AST-2406729.}
\vskip -6mm 

Being able to effectively read scientific plots, or chart understanding, is a central part toward building effective agents for science. However, existing multimodal large language models (MLLMs), especially open-source ones, are still falling behind with a typical success rate of 30\%-50\% on challenging benchmarks. Previous studies on fine-tuning MLLMs with synthetic charts are often restricted by their inadequate similarity to the real charts,
which could compromise model training and performance on complex real-world charts.
In this study, we show that modularizing chart generation and diversifying visual details improves chart understanding capabilities. 
In particular, we design a five-step data synthesis pipeline, where we separate data and function creation for single plot generation, condition the generation of later subplots on earlier ones for multi-subplot figures, 
visually diversify the generated figures, filter out low quality data, and finally generate the question-answer (QA) pairs with GPT-4o. 
This approach allows us to streamline the generation of fine-tuning datasets and introduce the effective chart dataset (ECD), which contains 10k+ chart images and 300k+ QA pairs, covering 25 topics and featuring 250+ chart type combinations with high visual complexity.
We show that ECD consistently improves the performance of various MLLMs on a range of real-world and synthetic test sets. Code, data and models are available at: \href{https://github.com/yuweiyang-anu/ECD}{https://github.com/yuweiyang-anu/ECD.}
\end{abstract}

\vspace{-4mm}
\section{Introduction}
\label{sec:intro}
\vskip -1mm

Chart understanding aims to enable MLLMs to accurately answer both descriptive questions (\eg, ``what is the label of the $x$-axis?'') that focus on recognizing basic chart elements, and reasoning questions (\eg, ``which city demonstrates the second-highest GDP growth rate?'') that require analytical capabilities. Several benchmarks extracted from real scientific papers, such as CharXiv \cite{wang2024charxiv}, have been developed to evaluate model performance on chart understanding tasks. But the performance of open-source models on these benchmarks remains mediocre, highlighting the need for more effective data and training approaches.

An advantage of chart data compared to natural images is that charts can be accurately synthesized using code, data, and standard libraries like Python's plotting packages \cite{matplotlib}. Unlike natural images which require extensive collection and annotation efforts and may suffer from low faithfulness when generated synthetically, charts can be created programmatically with high fidelity. This characteristic enables cost-efficient training data synthesis, an approach widely explored in the chart understanding field \cite{ xia2024chartx, he2024distill, meng2024chartassisstant, xu2023chartbench, han2023chartllama, xia2023structchart, kafle2018dvqa, kahou2017figureqa}.

% Researchers have increasingly focused on fine-tuning models with additional chart-specific training data to address these challenges, primarily through synthesizing specialized training sets. 
However, existing synthetic chart training datasets suffer from limited similarity to the real data. Early datasets such as PlotQA \cite{methani2020plotqa} and OpenCQA \cite{kantharaj2022opencqa}, while having some data variations, use fixed generation code templates with restricted modularity, resulting in only a few chart categories with limited style variations and inauthentic styles. ChartBench \cite{xu2023chartbench} has improved data diversity by generating data table separately 
but its visual diversity is limited by the codes for plotting the charts.
% but makes little change to codes. 
The recent ReachQA dataset \cite{he2024distill}, on the contrary, allows the code to be generated alongside the data table for the chart. However, this design limits the data patterns, further hurting the chart complexity. 
% changes the codes to promote visual diversity but has simple data patterns. Thus both exhibit limited visual complexity. 
%have improved diversity by incorporating random colors and plotting styles through GPT-4o, but still feature a limited range of chart types, simple chart data series, restricted subplot combinations, and relatively low visual complexity. 
These limitations compromise their similarity to and, therefore, effectiveness on real-world data. % such as CharXiv.

Our work considers the above limitations and introduces a data synthesis pipeline that generates high-quality charts and question-answer (QA) pairs to mimic real charts. The pipeline begins with human-defined chart functions and separated data generation, so that GPT-4o \cite{achiam2023gpt} can focus on generating more complex data. These data and functions are fed into Python to render a single plot. To create coherent multi-subplot charts, we use previous subplots as a condition when generating new subplots. We further diversify the charts by randomly adding visual elements such as annotations, area shadings, arrows, zoom-in insets, and subplot titles. Then, we rate the generated charts based on visual clarity and semantic coherence, filtering out low-quality examples before finally generating QA pairs using GPT-4o.

This pipeline allows us to create the effective chart dataset (ECD). It has over 10,000 images and 300,000+ QA pairs, belonging to 29 chart types and over 250 combinations of different chart types (Fig.~\ref{fig:motivation}). By fine-tuning open-source MLLMs with ECD, we show overall improvements on various test sets and that  % without directly training on existing QA benchmarks. 
our improvements surpass gains achieved with existing chart training sets.%, suggesting the potential for even greater enhancements with larger training sets. 
% Our analysis of question types reveals that descriptive questions are particularly valuable for strong MLLMs such as Phi-3-vision on the challenging CharXiv benchmark.

% \begin{itemize}[leftmargin=*]
%     \item[•] We introduce an effective chart data synthesis pipeline.
%     \item[•] We introduce the effective chart dataset that contains highly diverse chart and QA pairs for MLLM fine-tuning. 
%     \item[•] We achieve improved and competitive chart QA performance and derive interesting insights. 
% \end{itemize}

% refine the related work -> completed

\begin{figure*}[t]
  \centering
   \includegraphics[width=\linewidth]{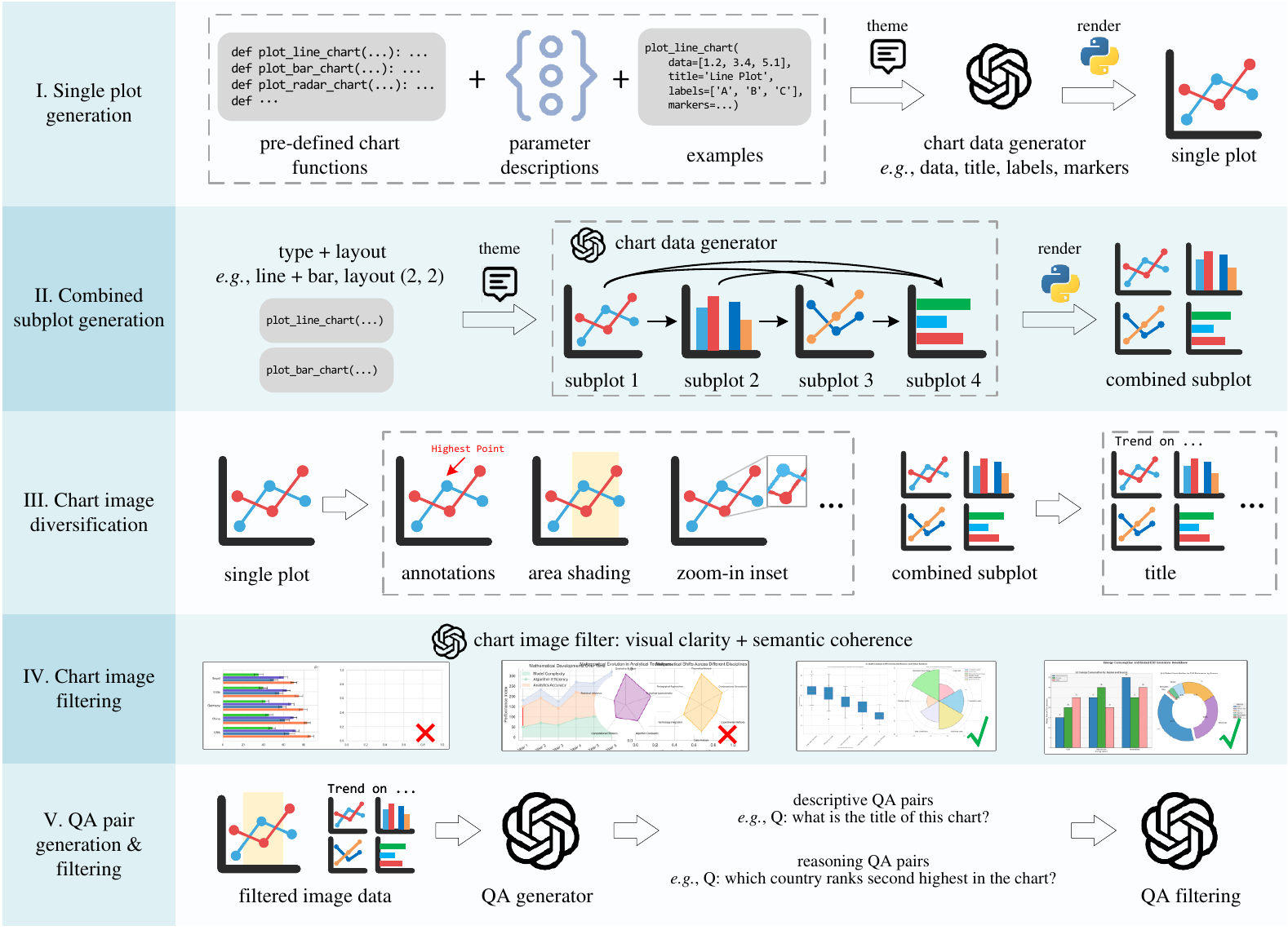}
    \vskip -2mm
   \caption{Our chart generation pipeline consists of five key steps: \textbf{I. Single plot generation} (Sec.~\ref{secsec:single}) creates individual charts using pre-defined chart functions and a separate chart data generator to produce data, titles, labels, and markers for each plot. \textbf{II. Combined subplot generation} (Sec.~\ref{secsec:multi}) produces multi-plot layouts through conditional sequential generation, where each subplot is informed by previous ones to maintain semantic coherence. \textbf{III. Chart image diversification} (Sec.~\ref{secsec:diversification}) enhances visual complexity and realism by adding annotations, area shading, zoom-in insets, and other stylistic variations including font modifications and axis border adjustments. \textbf{IV. Chart image filtering} (Sec.~\ref{secsec:filter}) uses rating metrics based on visual clarity and semantic coherence to filter out low-quality chart generations and ensure chart visual quality. \textbf{V. QA pair generation \& filtering} (Sec.~\ref{secsec:qa}) creates both descriptive questions and reasoning questions for the chart images, followed by additional quality filtering.}
   \label{fig:chart-synthesis-pipeline-all}
   \vskip -2mm
\end{figure*}

\section{Related Works}
\label{sec:related-works}

\textbf{Real-world chart datasets} %are collected from actual publications, websites, and other authentic sources. Datasets 
such as ChartQA \cite{masry2022chartqa}, OpenCQA \cite{kantharaj2022opencqa}, and CharXiv \cite{wang2024charxiv} extract charts from online sources, providing genuine examples of visualizations encountered in practice. Despite their authenticity, such real-world datasets  suffer from limited scale due to the effort required for collection and annotation. 

\textbf{Synthetic chart datasets} have gained popularity due to the programmatic nature of chart generation. Early datasets such as FigureQA \cite{kahou2017figureqa}, DVQA \cite{kafle2018dvqa}, PlotQA \cite{methani2020plotqa} and SimChart9k \cite{xia2023structchart} only generate basic chart types (\textit{e.g.}, line, pie, and bar charts) using template-based approaches and exhibit limited complexity. More recent synthetic datasets, including ChartLlama \cite{han2023chartllama}, ChartBench \cite{xu2023chartbench}, ChartAssistant \cite{meng2024chartassisstant}, NovaChart \cite{hu2024novachart} and ChartX \cite{xia2024chartx} have expanded the number of images, chart types, and styles, but still demonstrate insufficient visual complexity compared to authentic scientific visualizations. ReachQA \cite{he2024distill} uses large language models for code augmentation, but its data complexity and scale are limited. This work presents a new data synthesis pipeline and dataset, and extensive analysis demonstrates that our approach holistically address these limitations.

\textbf{Multimodal large language models for chart understanding} generally adopt two strategies. 
The first employs a two-stage approach: extracting charts into structured data tables or plotting codes, which then use additional modules for downstream question answering \cite{lee2023pix2struct,liu2023deplot,liu2023matcha,xia2023structchart,chen2024onechart}. 
The second strategy typically leverages chart-caption pairs for visual-language embedding alignment, followed by instruction tuning in an end-to-end manner \cite{han2023chartllama,liu2023mmc, cuarbune2024chart,meng2024chartassisstant, yang2024askchart, li2024synthesize,zhang2024tinychart,zeng2024advancing,masry2024chartinstruct,masry2024chartgemma, ye2023ureader,masry2023unichart, zadeh2024text2chart31,xu2024chartmoe,huang2024evochart}. Given the powerful and generalizable visual-language alignment of MLLMs, performing supervised fine-tuning (SFT) is equally effective. 

\textbf{Data synthesis for training MLLMs} is typically classified into two paradigms: (1) collecting real images and then synthesize instructions or (2) jointly synthesizing images and instructions. 
Some works, such as ProVision \cite{zhang2024provision}, LAMM \cite{yin2024lamm}, FM$^{2}$DS \cite{abaskohi2024fm2ds}, DSPT \cite{cheng2024domain} rely on publicly available image datasets or multimodal documents. 
Structured constructs (\textit{e.g.,} scene graphs) or GPT are then used to generate diverse instructions that emphasize fine-grained visual relationships.
To reduce image collection efforts and enhance privacy protection, a few automatic image-text pair generation methods \cite{liu2024synthvlm, aboutalebi2024magid, jiao2024img,brooks2023instructpix2pix} have been proposed.
% They create instructions using LLMs to guide the diffusion model in generating images, followed by further quality assessment and filtering. 
The aforementioned methods focus on generating natural images.
In contrast, programmable images like charts \cite{shi2024chartmimic,yang2024matplotagent}, SVGs \cite{xing2024empowering,rodriguez2023starvector}, \textit{etc.} exhibit different characteristics. These images can be fully represented in lossless formats like code, enabling a more precise synthesis of instructions. This paper therefore focuses on chart and QA pair synthesis to enhance MLLMs' chart comprehension capabilities.

\section{Proposed Data Synthesis Pipeline}

We aim to generate realistic plots whose visual content and stylistic characteristics are similar to 
% overlap with (and form a superset of) 
real-life scientific plots with the assistance from GPT-4o \cite{achiam2023gpt}. 
% \footnote{This project is supported by Microsoft's Accelerating Foundation Models Research program, and we are utilizing the resources that AFMR has provided, particularly the tokens for GPT-4o.} 
However, generating charts with diverse patterns of data while maintaining realism and accuracy of QA pairs presents challenges, and even GPT-4o itself demonstrates imperfections. % when reasoning about scientific charts.
To address this, we introduce a modular and systematic approach as illustrated in Fig.~\ref{fig:chart-synthesis-pipeline-all}, decomposing the synthesis process into five steps described in the subsections below.

\vspace{-0.5mm}
\subsection{Generating a Single Plot}
\label{secsec:single}

The foundation of our pipeline is to generate single plots with rich distributions. Instead of requesting GPT-4o to generate complete visualization codes alongside the data table, we strategically decompose the task by providing GPT with three key inputs: (1) a chart theme (\textit{e.g.}, mathematics, computer science, economics), (2) chart functions that can plot charts of specific types, and (3) parameter descriptions with few-shot examples. Given these inputs, GPT-4o generates appropriate data table and arguments including titles, legends, axis labels, \textit{etc.} 
% that collectively tell a coherent data story. 
By focusing on a step-by-step generation, we ensure the improvement on data distributions and the semantic connection between data values and corresponding textual elements. Specifically, we define 29 chart functions covering distinct chart types including line, bar, pie, scatter, and other common visualization formats. Each function contains modular Python code for rendering a chart based on specific parameters including data, title, axis labels, markers, line styles, \textit{etc.} We generated 10,875 single-plot figures through this process. 

% While these initial plots exhibit some visual variation through different line styles, marker types, and color schemes, their appearance diversity remains constrained by the function parameters. We address this limitation in Section~\ref{secsec:diversification} by introducing additional techniques to enhance visual complexity and stylistic diversity.

\subsection{Generating Combined Subplots}
\label{secsec:multi}

%Scientific visualizations frequently incorporate multiple subplots to present related information within a single figure. To accurately represent this common practice, o
%Our pipeline extends beyond individual plots to generate coherent multi-subplot visualizations that maintain semantic relationships across components.
To generate combined subplots, we implement a conditional generation approach: 
given specific chart types and layout configuration, it generates subplots iteratively, conditioned on prior subplots. 
% and conditions the later subplots on previous ones. 
This ensures consistency across the entire visualization. 
% that begins by . 
% We generate subplots iteratively, where each new subplot is conditioned on the data from all previously generated subplots. 
For example, as illustrated in Fig.~\ref{fig:chart-synthesis-pipeline-all}, when generating subplot 3 (a line chart),
our system incorporates data from both subplot 1 (another line chart) and subplot 2 (a bar chart) to maintain thematic coherence. This conditional generation strengthens the relationships between subplots, emulates human-designed scientific figures where multiple subplots present complementary data perspectives. Through this process, we generated 6,006 multi-subplot figures, with an average of 4 subplots per figure.

\setlength{\tabcolsep}{2.0mm}
\begin{table*}[t]
\centering
% \resizebox{\linewidth}{!}{
\small
\begin{tabular}{l|ccccccccc}  
\toprule
\multicolumn{1}{l|}{{Dataset}}          & {\#Themes}   & {\#Chart types}      &  \#Type Combo.        & {\#Images} & {Synthetic}       & {Subplots}      & {\#QA Pairs}         & {Training set}    
% & {Scalability}               
\\ \hline
%\multicolumn{8}{l}{\textit{Benchmark}} \\
\multicolumn{1}{l|}{OpenCQA \cite{kantharaj2022opencqa}}   & 10      & 5         &   1                                            & 1.2k                    & {{$\times$}} & {{$\times$}} & 1.2k  &  {$\times$} 
% & {{$\times$}} 
\\
\multicolumn{1}{l|}{ChartX \cite{xia2024chartx}}     & 22     & 18        &1                                                 & 6k                      & {\textbf{\checkmark}} & {\textbf{$\times$}} & 6k & {$\times$} 
% & {\textbf{\checkmark}} 
\\
\multicolumn{1}{l|}{CharXiv \cite{wang2024charxiv}}       & - &-                                       & -                      & 2.3k                    & {$\times$} & {\textbf{\checkmark}} & 5k   & {$\times$} &
% {\textbf{$\times$}} 
\\ \hline
%\multicolumn{8}{l}{\textit{Dataset}} \\
\multicolumn{1}{l|}{PlotQA \cite{methani2020plotqa}}      & -     & 3      & 1                                                   & 224k                    & {\textbf{\checkmark}} & {{$\times$}} & 29M  &  {\checkmark} & 
% {{$\times$}} 
\\
\multicolumn{1}{l|}{ChartQA \cite{masry2022chartqa}}     & 15      & 3       & 1                                                 & 21.9k                   & {{$\times$}} & {{$\times$}} & 32.7k  &  {\checkmark} & 
% {{$\times$}} 
\\
\multicolumn{1}{l|}{ChartLlama \cite{han2023chartllama}}   & 10    & 10    & 1                                                    & 11k                      & {\textbf{\checkmark}} & {\textbf{$\times$}} & 160k  &  {\checkmark} & 
% {\textbf{\checkmark}} 
\\

\multicolumn{1}{l|}{ChartBench \cite{xu2023chartbench}} & 9  & 9      &{4}                                                     & 68.7k                   & {\checkmark} & {{$\times$}} & 618.5k  & {\checkmark} 
% & {{$\times$}} 
\\
\multicolumn{1}{l|}{ChartAssistant \cite{meng2024chartassisstant}}     & -   & 9   &1                                                    & -                      & {\textbf{$\times$}} & {\textbf{$\times$}} & -  &  {\textbf{\checkmark}} & 
% {$\times$} 
\\
\multicolumn{1}{l|}{SimChart9k \cite{xia2023structchart}}  & -     & 3    &1                                                    & 9k                      & {\textbf{\checkmark}} & {\textbf{$\times$}} & 9k  &  {\checkmark} & 
% {$\times$} 
\\

\multicolumn{1}{l|}{MMC \cite{liu2023mmc}}  & 5   & 6          & -                                                & 600k                    & {\textbf{$\times$}} & {\checkmark} & 600k & {\checkmark} & 
% {$\times$} 
\\
\multicolumn{1}{l|}{ReachQA \cite{he2024distill}}  & -   & {10}      &-                                                  & 3.7k                    & {\checkmark} & {\checkmark} & 22k        & {\checkmark} & 
% {\checkmark} 
\\ 
% \midrule[0.1em]
\hline
\multicolumn{1}{l|}{ECD (ours)} & \textbf{25}  & \textbf{29} & \textbf{252}          & {10.5k}           & {\textbf{\checkmark}} & { \textbf{\checkmark}} & {321.5k}   & {\checkmark} & 
% {\textbf{\checkmark}} 
\\ 
% \bottomrule[0.15em]
\bottomrule
\end{tabular}
\vspace{-1mm}
\caption{Comparing the effective chart dataset (ECD) with existing chart datasets. ECD has the most themes, chart types, and type combinations. 
Although it is not the biggest, experiments show it enables consistent improvements on MLLM chart understanding capabilities. 
% Its training set consists of single plots and plot combinations and consistently improves MLLM chart understanding capabilities although its number of images and number of QA pairs are not the highest.
} 
\vspace{-1mm}
\label{tab:statistics}
% }
\end{table*}

\subsection{Chart Image Diversification}
\label{secsec:diversification}
Because of the use of human-defined chart functions in Section \ref{secsec:single} and Section
 \ref{secsec:multi}, the generated charts exhibit limited style diversity. To solve this problem, we implement a diversification process that modifies the Python code underlying each chart.
%Up to this stage, the generated charts, though have rich data table patterns and coherency across multiple subplots,  exhibit limited style diversity. This is because of the usage of predefined chart functions, where the visual diversity primarily comes from different chart function arguments. To enhance visual diversity and better represent the stylistic richness of real-world scientific visualizations, we implement a diversification process that modifies the Python code underlying each chart.
For single-plot figures, we prompt GPT-4o to alter the code by applying randomly selected visual enhancements from a predefined set of diversification strategies. These include operations such as: \textit{adding arrows}, \textit{annotations}, or \textit{highlights}; \textit{modifying font color}, \textit{style}, or \textit{size}; \textit{using gradient fills} or \textit{area shading}; \textit{removing axis borders}; and \textit{incorporating zoom-in insets} where appropriate.

For multi-subplot figures, we apply diversification after the subplot creation process (Section~\ref{secsec:multi}), modifying all subplots within one figure through a single GPT-4o request to ensure visual coherence. In addition to the enhancements used for single plots, we implement subplot-specific modifications such as incorporating overarching titles for the entire visualization. Additionally, we enable the use of supplementary visualization libraries beyond Matplotlib, such as Seaborn, to further enhance visual aesthetics. Moreover, we observe that some chart images, especially multi-subplot charts, have overly high resolutions or improper aspect ratios, resulting in unsuitable font sizes. To fix this, we add a post-processing step, prompting GPT-4o to adjust parameters like `figsize' or `dpi' in the code for better visualization. We apply diversification techniques and post-processing to all single-plot figures and multi-subplot figures and successfully generate 16,829 figures, substantially increasing the visual complexity and stylistic variation of ECD.

\subsection{Filtering Low-quality Charts}
\label{secsec:filter}

Despite our carefully designed pipeline, we cannot guarantee optimal visual quality for all generated charts. Common issues include excessive blank areas, overcrowded text elements, and misalignment between visual implementation and intended design. To address these challenges, we develop a quality filtering strategy that evaluates charts using two metrics: 1) \textbf{visual clarity} $r_\text{vis}\left(\bm{x}, c_\text{layout}\right)$, a function of the chart image $\bm{x}$ and the layout condition $c_\text{layout}$; 2) \textbf{semantic coherence} $r_\text{sem}\left(\bm{x}, c_\text{theme}\right)$, a function of the chart image $\bm{x}$ and the semantic condition $c_\text{theme}$. $r_\text{vis}$ measures how effectively a chart presents its data, considering readability, appropriate use of visual elements, absence of clutter, \textit{etc.} 
$r_\text{sem}$ assesses how well the chart elements and subplots maintain a consistent theme. Both metrics are evaluated using GPT-4o with specialized prompts (see supp. material). 

For each chart, we calculate a score by averaging its visual clarity and semantic coherence $r\left(\bm{x}, c_\text{layout}, c_\text{theme}\right) = \frac{\left(r_\text{s}+r_\text{v}\right)}{2}$, then compare this average to the mean score across the entire dataset $\bar{r}$.
% \begin{equation}
% I=\left \{ I_{i}\mid \frac{R_{i}^{v} + R_{i}^{s}}{2} >\frac{1}{N} \sum_{i=0}^{N} \frac{R_{i}^{v} + R_{i}^{s}}{2}  \right \},
% \end{equation}
% where $I$ represents the set of filtered chart plots, and $N$ denotes the total number of chart plots before filtering. 
We retain only charts whose quality score exceeds the dataset mean, \ie, $r\left(\bm{x}, c_\text{layout}, c_\text{theme}\right)>\bar{r}$. This method ensures our final dataset consists of charts that meet or exceed average quality. Starting with 16,829 figures, we filter out approximately 37.4\%, resulting in 10,535 high-quality chart images.

\begin{figure*}[t]
  \centering
   \includegraphics[width=0.9\linewidth]{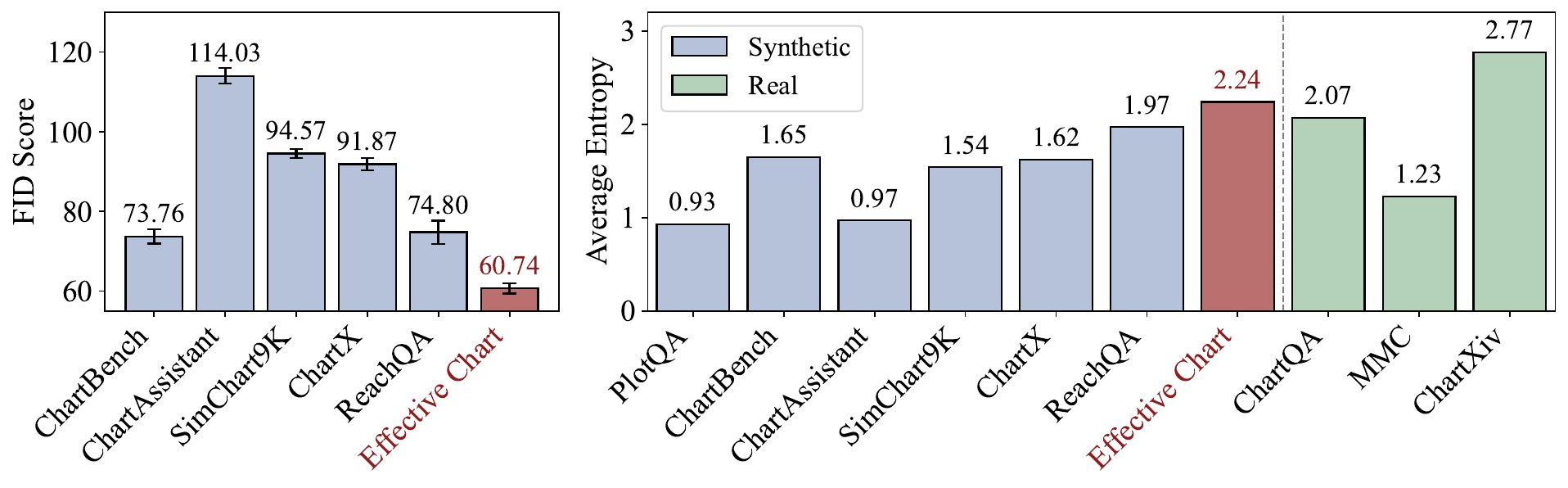}  
  \vskip -2mm
   \caption{Data complexity and realism comparison beween ECD and other datasets. 
   \textbf{Left}: FID comparison. FID is computed between each compared dataset with CharXiv~\cite{wang2024charxiv}. ECD has the lowest FID score so has stronger alignment with real scientific charts. \textbf{Right}: We measure complexity using average entropy of pixel values across images. ECD achieves higher average entropy than the synthetic datasets. 
   }
   \label{fig:dataset measurements}
   \vspace{-3mm}
\end{figure*}
\subsection{Synthesis and Filtering of QA Pairs}
\label{secsec:qa}

The final stage of our pipeline focuses on generating high-quality question-answer pairs that can effectively train multimodal models for chart understanding tasks. We use GPT-4o to synthesize these pairs by providing it with the filtered chart images, their generation code, and the underlying data. We categorize the generated QA pairs into descriptive and reasoning types. Descriptive questions focus on identifying and describing basic visual elements of the chart, whereas reasoning questions require analytical thinking and inference about the data relationships.
For descriptive questions, we retain only the final answer as the ground truth. For reasoning questions, we preserve both the step-by-step rationale and the final answer, enabling models to learn both the outcome and the reasoning process. 

To ensure quality, we implement a confidence-based filtering mechanism. When generating each QA pair, we request GPT-4o to assign a confidence score on a scale of 1 to 5, reflecting its certainty in the correctness of the answer. We only retain pairs with the highest confidence score of 5, eliminating potentially ambiguous or erroneous pairs.
Starting with 348,862 initially generated QA pairs from 10,535 images, we filtered out approximately 7.8\% based on the confidence scoring mechanism. The final dataset contains 321,544 high-quality QA pairs that cover both descriptive and reasoning tasks.

% @ting, instead of tsne plot, we calculate the FID score as the measurement.

\section{Effective Chart Dataset}

The above pipeline allows us to construct the effective chart dataset (ECD). %Our goal is to provide an open-source collection of high-quality, diverse, and complex scientific charts for future studies. 
Table \ref{tab:statistics} presents a comparative analysis between ECD and existing chart datasets.

\textbf{Styling coverage.} ECD has 25 themes  and 29 chart types, surpassing all other synthetic training datasets and  
providing broader contextual coverage across fields including economics, biology, physics, and sociology. It also has 252 combination types, \emph{e.g.}, bar + line and pie + line. This is significantly higher than most datasets that only have single plots, such as PlotQA. CharXiv, MMC, and ReachQA also have some types of combinations, but their counts are unspecified. In terms of scale, ECD contains 10,535 chart images and 321,544 question-answer pairs. 

\textbf{Data realism.} We compute Fréchet inception distance (FID)~\cite{salimans2016improved} between various datasets and CharXiv, a challenging real-world datasets crawled from arXiv papers. We show the results in Fig.~\ref{fig:dataset measurements} left. We clearly observe that ECD has higher similarity with CharXiv than the other synthetic datasets as measured by FID. 

\textbf{Data complexity.} We measure complexity using average entropy (see supp.), computed from image pixel values. For charts, average entropy essentially reflects the sophistication of individual charts, including the number of subplots, line intersections, data points, categorical elements, and the presence of non-trivial data patterns. We show complexity comparison of chart datasets in Fig.~\ref{fig:dataset measurements} right. We observe ECD has high average entropy among synthetic datasets. 

\textbf{ECDBench construction.} We construct an additional test set named ECDBench using the same five-stage data generation pipeline described earlier. This process produces single charts (\emph{e.g.}, line charts), multi-layout combination charts with one or two types (\emph{e.g.}, bar charts with a (3, 2) layout, bar + line charts with a (2, 4) layout), and three-type combination charts (\emph{e.g.}, bar + pie + area charts with a (1, 3) layout), resulting in 1,596 chart images after initial filtering.

To ensure data quality, we perform a two-stage rigorous review process. For the first stage, chart image filtering is conducted, where annotators remove images with significant visual recognition issues, yielding 1,224 charts. For minor issues such as overlaps, blank areas, incomplete legends or small fonts, they adjust the code and re-render the charts. In the second stage, annotators carefully verify the visual answerability of the questions by sampling one descriptive and one reasoning-based question from the candidate generated QA pairs for each chart. They also correct inconsistencies within the QA pairs and rewrite overly simplistic, inaccurate, or meaningless QA pairs based on their expertise. The aforementioned process ultimately produces 1,224 high-quality chart images and 2,448 high-quality QA pairs (each accompanied by one descriptive QA and one reasoning-based QA) for subsequent evaluation. More statistical information and evaluations can be found in the supplementary materials.

\begin{table*}[ht]
\centering
\small
\resizebox{\linewidth}{!}{
\begin{tabular}{l|cccc|cccccccccc}
\toprule
\multirow{2}{*}{{Models}} &  
\multicolumn{3}{c}{{CharXiv}} & 
{ChartQA} & 
\multicolumn{3}{c}{{ReachQA}} & 
\multicolumn{3}{c}{{ChartBench}} & 
{ChartX} &  
\multicolumn{3}{c}{{ECDBench}}\\
\cmidrule(lr){2-4} \cmidrule(lr){5-5} \cmidrule(lr){6-8} \cmidrule(lr){9-11} \cmidrule(lr){12-12} \cmidrule(lr){13-15}
& {Rea.} & {Des.}  & {Avg.} & {QA} & {Rea.} & {Recog.} & {Avg.} & {Binary} & {NQA} & {Avg.} &{QA} & {Rea.} & {Des.} & {Avg.} \\ 
\hline
\multicolumn{15}{c}{\textbf{Baselines}} \\ 
\hline
Human                 & 80.50 & 92.10 & 89.78 & -      & 65.10 & 84.60 & 74.85 &  - &  -  & -    & -    & -  & -  & -  \\
Random (GPT-4o) \cite{achiam2023gpt}        & 10.80 & 19.85 & 18.04 & 30.04  & 8.20  & 13.30 & 10.75 & -  & 22.73  & -  & 11.46 & 4.58 & 1.63 & 3.10 \\ 
\hline

\multicolumn{15}{c}{\textbf{Proprietary multimodal large language models}} \\ 
\midrule
GPT-4o mini \cite{achiam2023gpt}            & 34.10 & 74.92  & 66.76 & 77.52  & 27.20  & 53.50 & 40.35 & -  & 34.93  & - & 44.36 & 24.26 & 57.27 & 40.77 \\

GPT-4o \cite{achiam2023gpt}                  & 47.10 & 84.45  & 76.98 &  85.70   & 39.70  & 66.80  & 53.25 & -  & 52.88  & - & 58.33 & 35.62 & 70.18 & 52.90 \\
Claude 3.5 Sonnet \cite{anthropic2024claude3.5sonnet}       & 60.20  & 84.30  &  79.48  & 90.80   & 51.70  & 74.30  & 63.00  &  -  & 48.29 & - & 42.71 & 41.99 & 68.14 & 55.07 \\ 
\midrule

\multicolumn{15}{c}{\textbf{Open-source multimodal large language models}} \\ 
\hline
\rowcolor{gray!10}LLaVA-Next-Llama3-8B \cite{li2024llavanext}    & 19.70 & 38.90 & 35.06 & 64.56 & 8.00 & 23.30 & 15.65 & 61.18 & 38.29 & 58.64 & 27.69 & 4.74 & 17.16 & 10.95 \\
+ ECD   &  \cellcolor{green!8}{21.80} & \cellcolor{green!8}{59.05} & \cellcolor{green!8}{51.60} & \cellcolor{green!8}{68.64} & \cellcolor{green!8}{12.70} & \cellcolor{green!8}{37.50} & \cellcolor{green!8}{25.10} & \cellcolor{green!8}{64.21} & \cellcolor{green!8}{47.05} & \cellcolor{green!8}{62.30} & \cellcolor{green!8}{46.61} & \cellcolor{green!8}{16.50} & \cellcolor{green!8}{46.65} & \cellcolor{green!8}{31.58}  \\

\rowcolor{gray!10}MiniCPM-V2.6 \cite{yao2024minicpm}  & 25.40 & 52.65 & 47.20 & 76.96 & 14.50 & 48.10 & 31.30 & 75.67 & 52.33 & 73.08 & 47.31 & 15.15 & 39.95 & 27.53\\
+ ECD  &  \cellcolor{green!8}{26.90} & \cellcolor{green!8}{61.90} & \cellcolor{green!8}{54.91} & \cellcolor{green!8}77.28 & \cellcolor{green!8}17.50 & \cellcolor{green!8}48.40 & \cellcolor{green!8}32.95 & \cellcolor{red!8}70.77 & \cellcolor{green!8}52.81 & \cellcolor{red!8}68.77 & \cellcolor{green!8}50.43 & \cellcolor{green!8}18.14 & \cellcolor{green!8}52.21 & \cellcolor{green!8}35.17 \\
\rowcolor{gray!10}Phi-3-Vision \cite{abdin2024phi}   & 31.50  & 60.52 & 54.72 & 81.92 & 28.40 & 62.80 & 45.60 & 71.92 & 62.71 & 70.90 & 67.53 & 21.65 & 41.18 & 31.41   \\
+ ECD   & \cellcolor{green!8}{33.40} & \cellcolor{green!8}{68.00} & \cellcolor{green!8}{61.08} & \cellcolor{green!8}{84.88} & \cellcolor{green!8}{32.80} & \cellcolor{green!8}{64.50} & \cellcolor{green!8}{48.65} & \cellcolor{green!8}{73.49} &  \cellcolor{green!8}{63.76} & \cellcolor{green!8}{72.41} & \cellcolor{green!8}{71.44} & \cellcolor{green!8}{29.49} & \cellcolor{green!8}{59.31} & \cellcolor{green!8}{44.40} \\

\rowcolor{gray!10}Qwen2.5-VL-7B \cite{bai2025qwen2}   & 41.20  & 66.40 & 61.36 & 83.04 & 29.70 & 71.90 & 50.80 & 80.99 & 67.81 & 79.53 & 67.80 & 19.04 & 57.35 & 38.19   \\
+ ECD   & \cellcolor{red!8}{40.20} & \cellcolor{green!8}{74.20} & \cellcolor{green!8}{67.40} & \cellcolor{green!8}{85.32} & \cellcolor{green!8}{36.20} & \cellcolor{red!8}{70.80} & \cellcolor{green!8}{53.50} & \cellcolor{red!8}{79.35} &  \cellcolor{green!8}{70.86} & \cellcolor{red!8}{78.41} & \cellcolor{green!8}{70.83} & \cellcolor{green!8}{35.38} & \cellcolor{green!8}{66.34} & \cellcolor{green!8}{50.86} \\
\bottomrule
\end{tabular}
}
% }
\caption{
Comparing MLLMs on six test sets. 
``Rea.'', ``Des.'',  and ``Recog.'' mean reasoning,  descriptive, and recognition questions, respectively.
``Binary'' denotes ``yes/no'' answers, and NQA means numerical QAs. 
When fine-tuned on our effective chart dataset, except a few individual metrics (highlighted in 
\sethlcolor{red!8}
\hl{{red}}), the four open-source MLLMs display overall improvements (highlighted in 
\sethlcolor{green!8}
\hl{{green}}).
}
\label{tab:main}
\end{table*}

\section{Experiments}
\subsection{Datasets, MLLMs, and Evaluation Protocols}\label{secsec:datasets}
\textbf{Datasets.} To verify the effectiveness of the effective chart dataset (ECD), we test the trained models across six benchmarks: two real-world datasets (CharXiv \cite{wang2024charxiv} and ChartQA \cite{masry2022chartqa}) and four synthetic datasets (ReachQA \cite{he2024distill}, ChartBench \cite{xu2023chartbench}, ChartX \cite{xia2024chartx} and ECDBench), for a comprehensive evaluation. 

Specifically, for real-world data, the recently published \textbf{CharXiv} benchmark \cite{wang2024charxiv} comprises 2,323 challenging charts extracted from arXiv papers and includes 4,000 descriptive questions about basic chart elements and 1,000 reasoning questions requiring complex visual analysis. \textbf{ChartQA} \cite{masry2022chartqa} contains charts crawled from 4 online sources, with both human-annotated and machine-generated QA pairs, including a test set of 1,509 images with corresponding 2500 questions. Among the synthetic benchmarks, \textbf{ReachQA} \cite{masry2022chartqa} includes 500 chart images with 1,000 recognition and 1,000 reasoning questions; \textbf{ChartBench} \cite{xu2023chartbench} consists of 2,100 images paired with 18,900 QA pairs (2,100 numerical (NQA) and 16,800 binary ``yes/no'' questions); and \textbf{ChartX} \cite{xia2024chartx} contains 1,152 test chart images, each with a corresponding question-answer pair.

\textbf{MLLMs.} We comprehensively evaluate 4 open-source MLLMs: LLaVA-Next-Llama3-8B~\cite{li2024llavanext}, MiniCPM-V2.6~\cite{yao2024minicpm}, Phi-3-Vision~\cite{abdin2024phi} and the recent Qwen2.5-VL-7B~\cite{bai2025qwen2}, which have diverse architectures and training approaches. % within the open-source community. 
We fine-tune these models using ECD with a mix of parameter-efficient and full-parameter fine-tuning methods. We adopt LoRA~\cite{hu2022lora} for LLaVA-Next-Llama3-8B, MiniCPM-V2.6, and Qwen2.5-VL-7B, while performing full-parameter fine-tuning for Phi-3-Vision (4.2B). Specifically, we use a LoRA rank of 8, alpha of 16, a learning rate of $1\times 10^{-4}$, and train for 1 epoch for both LLaVA-Next-Llama3-8B and MiniCPM-V2.6, with batch sizes of 4 (on 4 A100 80G GPUs) and 1 (on a single A100 80G GPU), respectively. Phi-3-Vision is fine-tuned with a batch size of 96 and learning rates of $5\times 10^{-6}$ for most parameters and $5\times 10^{-8}$ for the projector, while Qwen2.5-VL-7B is fine-tuned with a LoRA rank of 64, alpha of 64, a batch size of 128, and learning rates of $1\times 10^{-4}$ for most parameters and $1\times 10^{-5}$ for the projector, both trained for 1 epoch on 4 A100 80G GPUs. In all experiments, the vision tower is frozen, and only the remaining parts of models are fine-tuned.

\textbf{Evaluation Protocols.} We perform QA tests on the official test sets without any specific additional prompts, \eg, chain-of-thought. By default, we use the widely accepted GPT-Acc~\cite{wang2024charxiv,he2024distill,xia2024chartx} as the accuracy metric, \ie, using GPT-4o to extract the answer and evaluate the correctness. For ChartQA, ChartBench NQA and ChartX, we follow the official guidelines, allowing a 5\% tolerance for questions with numerical answers. Our constructed ECDBench similarly follows the previously established evaluation pattern. For ReachQA and CharXiv, we directly utilize the official prompt templates for testing. Additionally, for the large number of ``yes/no'' binary questions in ChartBench, we employ a simple regular expression to match the answer and compute the accuracy.

\subsection{Main Evaluation}
\label{secsec:main-evaluation}

\textbf{Effective Chart consistently improves MLLMs in chart understanding.} We fine-tune the mentioned four open-source MLLMs with ECD and evaluate the fine-tuned models on six test sets. Results are summarized in Tab. \ref{tab:main}. We observe that ECD improves the overall performance of four MLLMs. For instance, LLaVA-Next-Llama3-8B's accuracy increases from 64.56\% to 68.64\% and from 27.69\% to 46.61\% on ChartQA and ChartX, respectively. On the challenging CharXiv benchmark, ECD boosts Phi-3-Vision's accuracy from 31.50\% and 60.52\% to 33.40\% and 68.00\% for reasoning and descriptive questions.

ECD fine-tuning causes slight performance drops in specific metrics, as shown by the red-highlighted results in Tab.~\ref{tab:main}. MiniCPM-V2.6 and Qwen2.5-VL-7B see decreases on ``Binary'' questions in ChartBench, likely due to its unique distribution with ``yes/no'' questions only. For the latest SOTA Qwen2.5-VL-7B, while it achieves notable average improvements on CharXiv and ReachQA, minor drops (\textit{e.g.}, reasoning accuracy on CharXiv, from 41.20\% to 40.20\%, recognition accuracy on ReachQA, from 71.90\% to 70.80\%) may be attributed to its pre-training on clean chart data, which closely resembles certain benchmarks. Moreover, the data mixture ratio during fine-tuning might not have been fully optimized for specific tasks or models.

On ECDBench, performance significantly improves due to the similar QA distribution advantage: LLaVA-Next-Llama3-8B's average accuracy increases from 10.95\% to 31.58\%, and Qwen2.5-VL-7B achieves the highest accuracy, rising from 38.19\% to 50.86\%.

\begin{figure*}[t]
  \centering
  \includegraphics[width=0.98\linewidth]{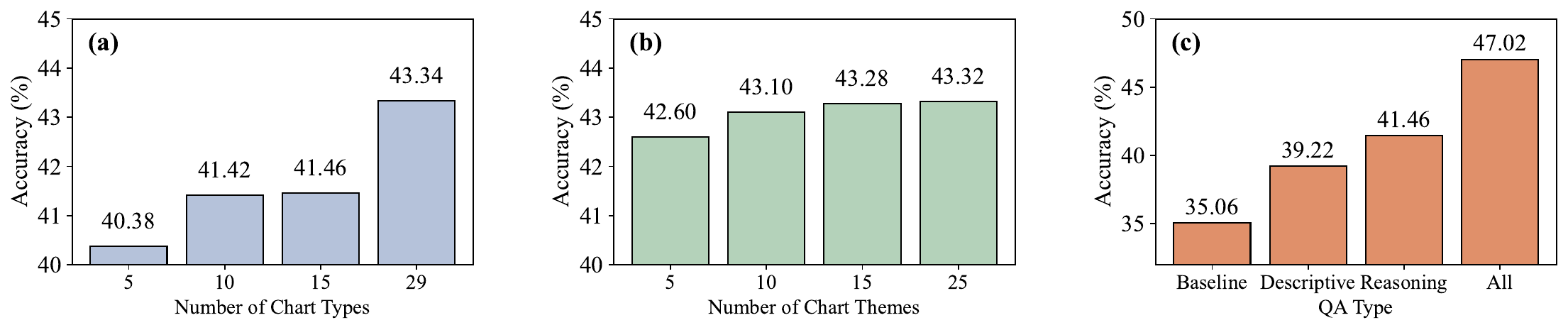}
   \vspace{-3mm}
   \caption{Impact of the number of chart types (a), chart themes (b), and QA type (c). Average accuracy on CharXiv is reported.
   In the compared variants, we make sure the numbers of images are the same. LLaVA-Next-Llama3-8B is selected as the baseline model. }
   \label{fig:ablation_studies}
    \vspace{-2mm}
\end{figure*}

\begin{table}[t]
\centering
\scriptsize
\setlength{\tabcolsep}{1.0mm}
{
\begin{tabular}{l|cccccc}
\toprule
\multicolumn{1}{c|}{\multirow{2}{*}{Training set}} & \multicolumn{6}{c}{{Test set}}                                                                                 \\ \cline{2-7} 
\multicolumn{1}{l|}{}                       & {CharXiv}                 & {ChartQA}               & {ReachQA}               & {ChartBench}             & {ChartX}  & {ECDBench}  \\ \hline
\rowcolor{gray!10}Baseline           & 35.06                   & 64.56                 & 15.65                & 58.64                  & 27.69  & 10.95             \\ \hline
ChartQA                       & \cellcolor{green!8}{\ul 35.16}           & \cellcolor{green!8}\textbf{68.92} & \cellcolor{red!8}15.00          & \cellcolor{green!8}{61.41}           & \cellcolor{green!8}31.51  & \cellcolor{green!8}13.11        \\
ChartBench                     & \cellcolor{red!8}32.86           & \cellcolor{red!8}61.56         & \cellcolor{green!8}18.35          & \cellcolor{green!8}\textbf{75.13} & \cellcolor{green!8}37.33  & \cellcolor{green!8}10.99         \\
ReachQA                      & \cellcolor{red!8}30.68           & \cellcolor{red!8}64.50         & \cellcolor{green!8}{\ul 24.35}          & \cellcolor{red!8}57.59           & \cellcolor{green!8}{\ul 39.24} & \cellcolor{green!8}{\ul 13.48}          \\
ECD                          & \cellcolor{green!8}\textbf{51.60} & \cellcolor{green!8}{\ul 68.64}   & \cellcolor{green!8}\textbf{25.10} & \cellcolor{green!8}{\ul 62.30}     & \cellcolor{green!8}\textbf{46.61} & \cellcolor{green!8}\textbf{31.58} \\ \bottomrule
\end{tabular}
}
\vspace{-1mm}
\caption{
% Performance gains from different training sets. 
Comparing different chart training sets of their fine-tuning performance on 6 test sets. 
%Performance improvements from synthetic training datasets. 
\sethlcolor{green!8}
\hl{{Green}} indicates performance growth and \sethlcolor{red!8}
\hl{{red}} highlights decline. We use LLaVA-Next-Llama3-8B~\cite{li2024llavanext} as the baseline and fine-tune it on ChartQA~\cite{masry2022chartqa}, ChartBench~\cite{xu2023chartbench}, ReachQA~\cite{he2024distill}, and ECD. For each column, best number is in \textbf{bold}, and second best \underline{underlined}.
}
\label{tab:training comparison}
\vspace{-5mm}
\end{table}

\textbf{Comparison with other training datasets.} We then compare ECD with other chart understanding training sets, including ChartQA~\cite{masry2022chartqa}, ChartBench~\cite{xu2023chartbench}, and ReachQA~\cite{he2024distill}. To do so, we use these datasets to fine-tune LLaVA-Next-Llama3-8B, evaluate the fine-tuned models on six test sets, and report the average accuracy. Results are presented in Tab. \ref{tab:training comparison}. 

While the existing training sets improve on their own test sets and those with a similar distribution, for other test sets, the fine-tuned model often results in performance drops. For example, since ChartBench has a very different QA distribution, comprising mostly binary QA, training on it gives the best evaluation result for this dataset. However, it also falls behind on two other benchmarks.  

In comparison, ECD consistently improves the baseline performance across 6 test sets, while other chart training datasets show the mixed and less consistent results. In addition, ECD yields the largest improvement on the CharXiv, ReachQA, ChartX, and our ECDBench, and the second-largest improvement on ChartQA and ChartBench.

\subsection{Ablation Study}

To isolate the effects of specific design choices, we conduct a series of ablation studies. When comparing variants of ECD in this section, we maintain consistent image counts for fair comparisons. We select LLaVA-Next-Llama3-8B~\cite{li2024llavanext} as the base model for our study and conduct experiments on each plot image by randomly selecting five QA pairs (including both descriptive and reasoning QA pairs) to examine the chart types, themes, and QA type.

\textbf{Impact of number of chart types.} As shown in Fig.~\ref{fig:ablation_studies} (a), increasing the number of chart types from 5 to 29 yields progressive improvements on the CharXiv test set. This confirms the importance of diverse training chart types for effective generalization to real-world charts.

\textbf{Impact of number of chart themes.} Fig.~\ref{fig:ablation_studies} 
(b) demonstrates that expanding thematic coverage improves model performance, though the improvement gradually slows as the number of themes increases, with the best results achieved at 25 themes. This indicates the usefulness of having more themes.

\textbf{Impact of QA Type.} Fig.~\ref{fig:ablation_studies} (c) compares the pre-trained LLaVA-Next baseline and with the same fine-tuned with three ECD variants, including one with descriptive questions only, one with reasoning questions only, and the full ECD. We observe that using either descriptive (+4.16\%) or reasoning questions (+6.40\%) improves performance, with greater gains observed when using reasoning questions exclusively. The best model is obtained with the full ECD fine-tuning, which yields 47.02\% accuracy. Results validate the inclusion of both types of questions. 

\subsection{Further Analysis}
\textbf{Impact of data scale.} Experimental results on data scaling are shown in Tab.~\ref{tab:impact-of-data-scale}. We find that increasing data size (default scale is 10k) consistently improves average performance on the synthetic benchmark ReachQA. For real-world benchmark CharXiv, the gains saturate beyond 20k, probably due to the complexity of this task. As future work, we will scale up data in the pre-training phase. 

\vspace{-0.7em}
\begin{table}[ht]
\centering
\small
\setlength{\tabcolsep}{5.5pt}
\begin{tabular}{l|cccccc}
\toprule
{{Test set}} & 2k & 5k & 10k & 20k & 30k & 40k \\
\midrule
CharXiv        & 40.46 & 45.66 &  46.84    &  48.90  &  47.84  &    48.26   \\
ReachQA        & 18.25 & 18.85 &  20.85  &  22.25  &  23.50   &  24.75 \\
\bottomrule
\end{tabular}
% }
\vspace{-2mm}
\caption{Impact of data scale (5 selected QA pairs per image).}
\label{tab:impact-of-data-scale}
\end{table}

\vspace{-3mm}
\textbf{Impact of data mixture ratios.}
 We tried different ratios of descriptive and reasoning questions of ECD (50k QA pairs in total). Results on our CharXiv are shown in Tab.~\ref{tab:mixture-ratio}. Interestingly, we find the average accuracy metrics peak at a 2:3 or 1:1 ratio. This may be because our definition of descriptive/reasoning does not perfectly match the definition on the benchmarks. Another possible reason is the benefit of having both types of questions on VLM training, which needs further exploration. 

\vspace{-0.7em}
\begin{table}[ht]
\centering
\small
\setlength{\tabcolsep}{2.2pt}
\begin{tabular}{l|ccccccc}
\toprule
{{CharXiv}} & 0:5 & 1:4 & 2:3 & 1:1 & 3:2 & 4:1 & 5:0 \\
\midrule
Descriptive Acc & 51.25 & 53.88 & 53.75 & 54.90 & 54.25 & 52.92 & 47.48 \\
Reasoning Acc  & 20.20 &  19.20 & 20.90 & 20.50 & 20.00 & 20.90 & 22.50  \\
\rowcolor{green!10}
Average Acc & 45.04 & 46.94 & 47.18 & 48.02 & 47.40 & 46.52 & 42.48  \\
\bottomrule
\end{tabular}
% }
\vspace{-2mm}
\caption{Impact of data ratios (descriptive vs. reasoning).}
\vspace{-1mm}
\label{tab:mixture-ratio}
\end{table}

\vspace{-2mm}
\textbf{Impact of visual diversification.} We use FID (with CharXiv) and average pixel entropy to evaluate image quality with or without the diversification step (Section~\ref{secsec:diversification}). Results are presented in Tab.~\ref{tab:compaision-of-different-strategries} (a) and (b). It is shown that adding diversification is very beneficial, which decreases FID by 19.64 and increases average entropy by 0.57. 

% Need further checking ...
\begin{table}[]
\centering
\small
\begin{tabular}{l|cc}
\toprule
\multicolumn{1}{c|}{{Data Strategy}}  & \multicolumn{1}{c}{{FID$\downarrow$}} & \multicolumn{1}{c}{{Avg. Entropy$\uparrow$}} \\ \hline
(a) w/o Visual Diversification &      80.38 ± 1.45
                 &    1.67                   \\
(b) w Visual Diversification   &     \textbf{60.74 ± 1.32}                    & \textbf{2.24}                     \\ \hline
(c) w/o Image Filtering        &   66.20 ± 1.11                      &   2.16                    \\
(d) w Image Filtering          &  \textbf{60.74 ± 1.32}                      &  \textbf{2.24}                    \\ \bottomrule
\end{tabular}
\vspace{-2mm}
\caption{Comparing data processing strategies. We compare visual diversification and the image filtering strategies with not applying them. We report FID (with CharXiv) and average pixel entropy.}
\vspace{-6mm}
\label{tab:compaision-of-different-strategries}
\end{table}

\textbf{Importance of image and QA quality filtering.} In Tab.~\ref{tab:compaision-of-different-strategries} (c) and (d), we compare dataset quality with or without image quality filtering (Section \ref{secsec:filter}). If we remove quality filtering, the FID score with the CharXiv dataset increases by 5.46, and the average entropy decreases by 0.08. The results show that filtering out low-quality images is beneficial. 
Moreover, to validate the effectiveness of QA filtering (Section ~\ref{secsec:qa}), we remove the QA filtering step and compare the fine-tuned LLaVA-Next-Llama3-8B. 
% We make sure the number of QA pairs is the same, so we can evaluate the impact of filtering out low-quality QAs. 
Results are shown in Tab.~\ref{tab:qa-filtering-strategy}. We observe that ECD without QA filtering still improves the baseline by 11.58\% and that with QA filtering the improvement becomes 11.96\%. The additional improvement outlines the benefit of doing so. 

\linespread{0.87}
\vspace{-3mm}
\begin{table}[tbh]
\centering
\small
\setlength{\tabcolsep}{10.8pt}
\begin{tabular}{l|ccc}
\toprule
                     & \multicolumn{3}{c}{\textbf{CharXiv}}                                                                           \\ \cline{2-4} 
\multirow{-2}{*}{}   & Rea.                      &  Des.                      & Avg.                                   \\ \hline
LLaVA-Next-Llama3-8B & 19.70 & 38.90 & 35.06          \\
w/o QA Filtering     & 19.50 & 53.42 & 46.64  \\
w QA Filtering       & \textbf{20.90} & \textbf{53.55} & \textbf{47.02}  \\ \bottomrule
\end{tabular}

\caption{Quantitative performance comparison before and after QA filtering. The average performance improves after filtering.}
\label{tab:qa-filtering-strategy}
\vspace{-2mm}
\end{table}
\linespread{1.0}

\vspace{-4mm}
\section{Conclusion}

This paper introduces a chart synthesis pipeline and the effective chart dataset (ECD), a high-quality multimodal training resource that enhances chart understanding capabilities in existing MLLMs. Through systematic ablation studies and comparative evaluations, we demonstrate how each component of our pipeline contributes to the overall effectiveness of the resulting dataset.
The effective chart dataset generally improves the accuracy of MLLMs across multiple test sets.  
% and demonstrates excellent scaling properties, with performance gains continuing as dataset size increases. 
Particularly notable is the improvement on challenging real-world chart benchmarks like CharXiv, highlighting our approach's ability to bridge the gap between synthetic training data and authentic scientific visualizations.
%Our work provides both practical resources and methodological insights for the research community. 
The modular nature of our pipeline enables further extensions and customizations, while our findings regarding the importance of quality and complexity in synthetic data generation have broader implications for multimodal training dataset development.

{
    \small
    \bibliographystyle{ieeenat_fullname}
    \bibliography{main}
}

\clearpage
\maketitlesupplementary
\appendix

\section{Chart Types and Subplot Combinations}
\begin{figure}[bh]
  \centering
   \includegraphics[width=1.0\linewidth]{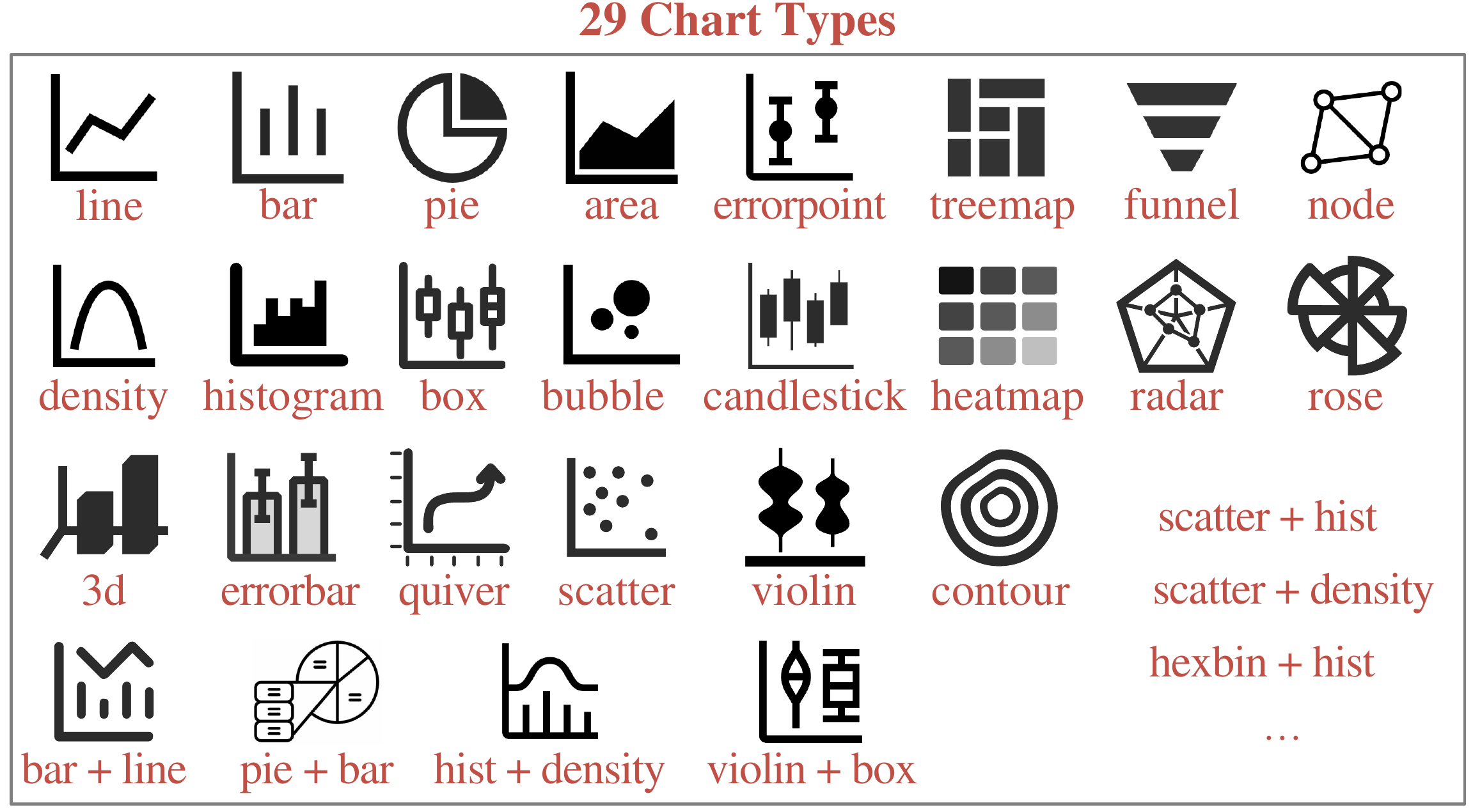}
   \caption{Distribution of Chart Types in our proposed effective chart dataset (ECD). Our dataset encompasses 29 diverse chart types, surpassing the variety present in existing datasets.}
   \label{fig:supple-chart-types}
\end{figure}

As illustrated in Fig.~\ref{fig:supple-chart-types}, our ECD dataset includes 29 distinct chart types for single plot, \textit{i.e.,} ``line'', ``bar'', ``pie'', ``area'', ``errorpoint'', ``treemap'', ``funnel'', ``node'', ``density'', ``histogram'', ``box'', ``bubble'', ``candlestick'', ``heatmap'', ``radar'', ``rose'', ``3d'', ``errorbar'', ``quiver'', ``scatter'', ``violin'', ``contour'', ``bar + line overlay'', ``pie + bar overlay'', ``histogram + density overlay'', ``violin + box overlay'', ``scatter + histogram overlay'', ``scatter + density overlay'' and ``hexbin + hist overlay''.

Additionally, for the combinations of subplots, our dataset only uses random combinations of 2 single-plot chart types (excluding overlays, can be the same type, \textit{e.g.,} ``line + line''). Following the pipeline outlined in this paper, it can be seamlessly extended to incorporate combinations of more chart types in practice, such as combinations involving 3 chart types, (\textit{\textit{e.g.,}} ``line + bar + pie'', \textit{etc.}). Ultimately, 252 distinct combinations are successfully generated (excluding unsuccessful examples after filtering).

\section{Chart Themes and Layout Distribution}
\begin{figure*}[t]
  \centering
   \includegraphics[width=0.9\linewidth]{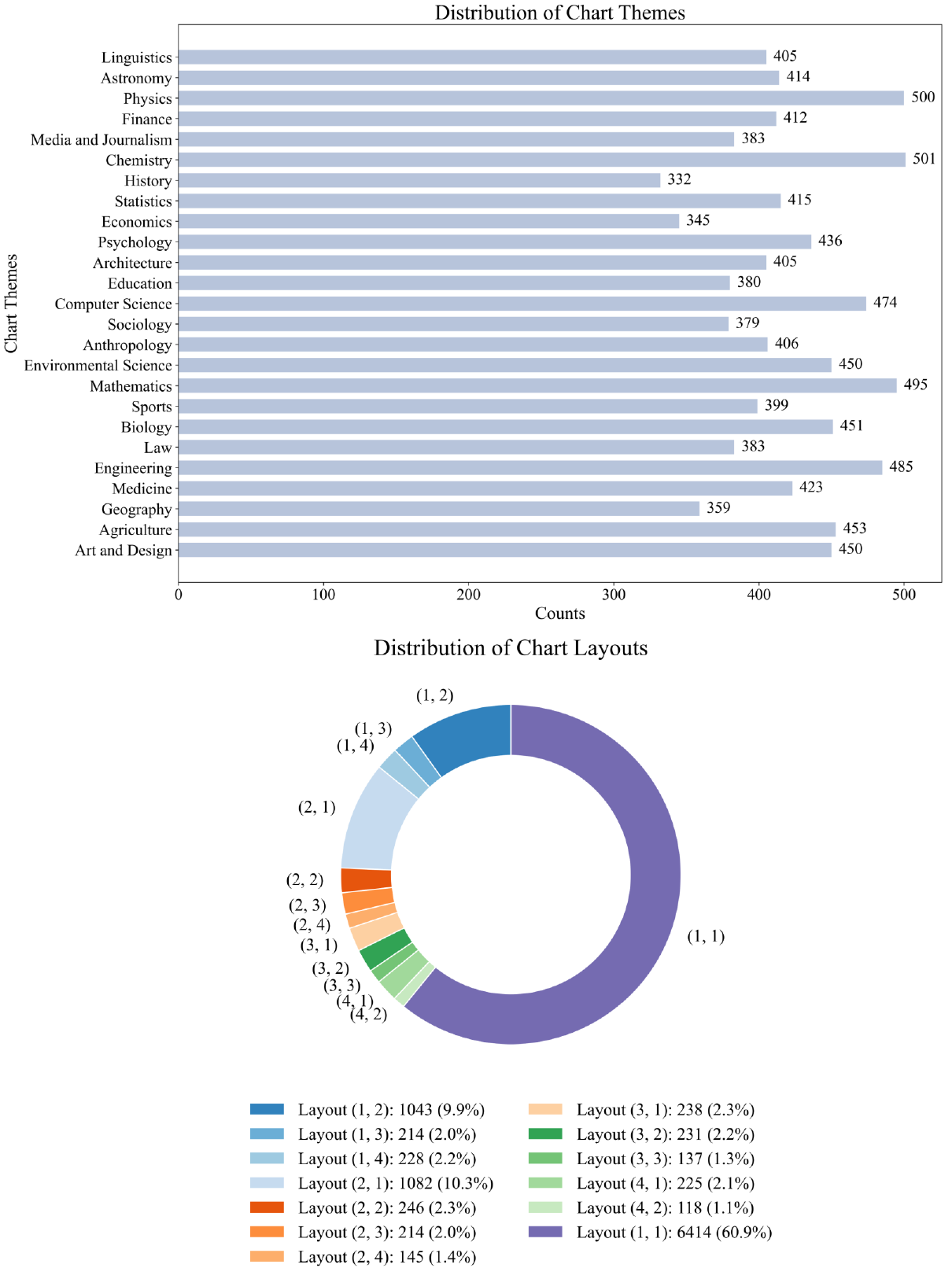}
   \caption{Distribution of themes and layouts in our proposed effective chart dataset (ECD). Our dataset comprises 25 distinct themes and 12 multi-plot layouts. It includes 6,414 single-chart images and 4,121 combined subplot images. Among the multi-plot layouts, 631 subplots consist of combinations of the same chart type, while 3,490 subplots involve combinations of different chart types. Following the pipeline proposed in this paper, our dataset can be further expanded to include more chart types and layouts.}
   \label{fig:supple-analysis_of_our_dataset}
\end{figure*}

As shown in Fig.~\ref{fig:supple-analysis_of_our_dataset}, we present the distribution of chart themes and layouts in the proposed ECD dataset. 

\textbf{Chart Themes Distribution.} Our dataset contains 25 scientific themes, with an average of approximately 421.4 images per theme.
The number of images in each theme is as follows: Art and Design: 450, Agriculture: 453, Geography: 359, Medicine: 423, Engineering: 485, Law: 383, Biology: 451, Sports: 399, Mathematics: 495, Environmental Science: 450, Anthropology: 406, Sociology: 379, Computer Science: 474, Education: 380, Architecture: 405, Psychology: 436, Economics: 345, Statistics: 415, History: 332, Chemistry: 501, Media and Journalism: 383, Finance: 412, Physics: 500, Astronomy: 414, Linguistics: 405. Among them, the Chemistry theme has the highest count, while the History theme has the lowest.

\textbf{Chart Layouts Distribution.} We include 13 primary layouts, ranging from ``1 by 1 (1, 1)'' to ``4 by 2 (4, 2)''. The "1 by 1" layout represents a single chart image, with a total of 6,414 instances (accounting for 60.9\%), while the other layouts represent multiple images, comprising 4,121 instances (accounting for 39.1\%). This includes 631 instances of multiple images of the same chart type (\textit{e.g.,} line + line subplots) and 3,490 instances of multiple images of different chart types (\textit{e.g.,} line + bar subplots). 

\begin{table}[h!]
\centering
\small 
\setlength{\tabcolsep}{4pt}  
\begin{tabular*}{\columnwidth}{@{\extracolsep{\fill}}lccc}
\hline
\centering Models & \multicolumn{3}{c}{ECDBench} \\
\cline{2-4}
                          & Reasoning & Descriptive & Average \\
\hline
\multicolumn{4}{c}{Claude Series} \\
Claude 3.5 Sonnet \cite{anthropic2024claude3.5sonnet}         & 41.99 & 68.14 & 55.07 \\
Claude 3.7 Sonnet \cite{anthropic2025claude3.7sonnet}         & 43.38 & \textbf{69.61} & 56.50 \\
Claude 4 Sonnet \cite{anthropic2025claude4sonnet}         & \textbf{44.20} & 69.36 & \textbf{56.78} \\
\hline
\multicolumn{4}{c}{Gemini Series} \\
Gemini-2.5-Pro \cite{comanici2025gemini}           & \textbf{44.36} & \textbf{76.88} & \textbf{60.62} \\
\hline
\multicolumn{4}{c}{GPT Series} \\
Random (GPT-4o) \cite{achiam2023gpt}          & 4.58 & 1.63 & 3.10 \\
GPT-4o-mini \cite{achiam2023gpt}               & 24.26 & 57.27 & 40.77 \\
GPT-4o \cite{achiam2023gpt}                     & 35.62 & 70.18 & 52.90 \\
o1 \cite{jaech2024openai}                      & 40.52 & 74.18 & 57.35 \\
o3 \cite{openai2025o3o4mini}                    & 56.13 & 74.51 & 65.32 \\
o4-mini \cite{openai2025o3o4mini}         & \textbf{57.03} & \textbf{77.45} & \textbf{67.24} \\
\hline
\multicolumn{4}{c}{Qwen-VL Series} \\
Qwen2.5-VL-7B \cite{bai2025qwen2}              & 19.04 & 57.35 & 38.19 \\ 
Qwen2.5-VL-32B \cite{bai2025qwen2}           & 24.92 & 53.92 & 39.42 \\
Qwen2.5-VL-72B \cite{bai2025qwen2}           & \textbf{38.81} & \textbf{68.46} & \textbf{53.64} \\
\hline
\end{tabular*}
\caption{Performance comparison of several closed-source and open-source MLLMs on ECDBench. Among the listed models, o4-mini achieved the best performance.}
\label{tab:comparison_on_ECDBench}
\end{table}

\section{ECDBench Statistics and More Results}
The ECDBench we constructed comprises a total of 1,224 chart images, including 364 single-plot charts and 860 combined subplot charts. Among the combined-subplot charts, 457 feature combinations of two chart types (\textit{e.g.,} ``bar + line''. with various layouts ranging from (1, 2) to (3, 3) / (4, 2), \textit{etc.}), while 403 involve combinations of three chart types (\textit{e.g.,} ``bar + pie + scatter''. with (1, 3) or (3, 1) layout). The average size of images in ECDBench is 1378 × 968 px. ECDBench includes 2,448 QA pairs, with each image accompanied by 1 descriptive QA and 1 reasoning-based QA. For the manual checks and annotations of ECDBench, we employ PhD students who are familiar with common scientific charts.

Tab.~\ref{tab:comparison_on_ECDBench} presents a more detailed comparison of evaluation results on several state-of-the-art (SOTA) proprietary and open-source MLLMs, including the Claude series, Gemini series, GPT series, and Qwen-VL series. Among all the evaluated MLLMs, o4-mini consistently achieves the highest performance across all three metrics (57.03\% reasoning acc, 77.45\% descriptive acc and 67.24\% average acc).

\section{FID and Avearage Pixel Entropy}
In this paper, we employ the FID (Fréchet Inception Distance) and Average Pixel Entropy as metrics to assess the image realism and complexity of chart datasets respectively. Here, we present the specific calculation formula.

% \textbf{IS Score:} Let the dataset be defined as a collection of chart images $I = \{x_1, x_2, ..., x_N\}$, where $N$ denotes the total number of images and each $x_i \in I$ represents an individual chart image. The IS Score can be formally expressed as:
% \begin{equation}
% \text{IS}(I) = \exp\left(\mathbb{E}_{x \in I} \left[\text{KL}\left(p(y|x) \parallel p(y)\right)\right]\right),
% \end{equation}
% where $p(y|x)$ denotes class probability distribution conditioned on image $x$ (from the inception model), $p(y) = \mathbb{E}_{x \in I}[p(y|x)]$ represents marginal class distribution over the entire dataset, $\text{KL}(\cdot\parallel\cdot)$ is Kullback-Leibler divergence. 

\textbf{Fréchet Inception Distance (FID):} 
For the real dataset \( I_{\text{real}} \) and Synthetic dataset \( I_{\text{syn}} \):
\begin{equation}
    \scalebox{0.8}{$
    \text{FID}(I_{\text{real}}, I_{\text{syn}}) = \|\mu_{\text{real}} - \mu_{\text{syn}}\|^2 + \text{Tr}\left(\Sigma_{\text{real}} + \Sigma_{\text{syn}} - 2\sqrt{\Sigma_{\text{real}}\Sigma_{\text{syn}}}\right)
    $},
\end{equation}
where \( \mu_{\text{real}} \) and \( \mu_{\text{syn}} \) are the mean feature vectors of the Inception network embeddings, \( \Sigma_{\text{real}} \) and \( \Sigma_{\text{syn}} \) are their covariance matrices, and \( \text{Tr}(\cdot) \) denotes the matrix trace operator.

\textbf{Average Pixel Entropy:} The average pixel entropy can be defined as:
\begin{equation}
H(x) = -\sum_{i=1}^{n} p(i) \log p(i),
\end{equation}
\begin{equation}
\text{AvgEnt}(I) = \frac{1}{N} \sum_{k=1}^{N} H(x_k),
\end{equation}
where $p(i)$ is the normalized histogram probability of intensity value $i$, $n$ is the number of discrete intensity levels (\textit{e.g.,} 256 for 8-bit images), $H(x_k)$ is entropy of the $k$-th image. 

\section{Computational and GPT-4o Cost} 
\begin{table}[ht]
\centering
\small
\setlength{\tabcolsep}{2.2pt}  
\renewcommand{\arraystretch}{1.1}  
\begin{tabular}{l|ccccc}
\toprule
{Dataset} & {\#Images} & {\#QA Pairs} & {\$Per-image} & {\$Per-QA} & {\$Total} \\
\midrule
CharXiv         & 2.3k   & 5k       & 0.37  & 0.170 & $\sim$0.85k \\
ReachQA         & 3.2k   & 19.9k    & 0.10  & 0.016 & $\sim$0.31k \\
ECD & 10.5k  & 321.5k   & 0.20  & 0.007 & $\sim$2.15k  \\
\bottomrule
\end{tabular}
\caption{Cost (\$) comparison between ECD and existing datasets.}
\label{tab:cost-comparison-synthetic}
\end{table}

Generating 1,000 single-plot images takes an average of 7.1 seconds per image and 22.40 USD in total. On the other hand, generating 1,000 combined subplots costs 22.6 seconds per image and 85.12 USD in total. During diversification, each set of 1,000 images costs 30.14 USD and requires an average of 40.18 seconds per image. The figure size post-processing and the image filtering costs 27.90 USD per 1,000 images in total. Additionally, generating QA pairs for 1,000 chart images costs 38.97 USD. In total, generating ECD costs $\sim$2,145 USD. We add cost comparison in Tab.~\ref{tab:cost-comparison-synthetic}. While ECD has the highest total cost due to its large number of image and QA pairs, its per QA cost is the lowest, and its per-image cost is medium (but not rediculously high on 0.2\$ per image), because it has many multi-subplot charts, which is essential in its design. We also tried replacing GPT-4o with Qwen2.5-VL-32B for training data generation. Results are shown in Tab.~\ref{tab:ecd-qwen2.5-vl}. We find that both yield improvement on ReachQA benchmark, but the dataset generated by Qwen2.5-VL-32B does not improve performance on CharXiv. 

\begin{table}[ht]
\centering
\small
\setlength{\tabcolsep}{10.5pt}
\begin{tabular}{l|cc}
\toprule
{Models} & {CharXiv} & {ReachQA} \\
\midrule
LLaVA-Next-Llama3-8B       &    35.06   &  15.65     \\
+ ECD by Qwen2.5-VL-32B    &  34.90  &  21.40    \\
+ ECD by GPT-4o     &   45.66  & 18.85       \\
\bottomrule
\end{tabular}
% }
\caption{Comparison on ECD generation (5k images, 30k QAs).}
\label{tab:ecd-qwen2.5-vl}
\end{table}

\section{Cross-task MLLM SFT Impact} To assess the impact of our synthetic chart QA data on model generalization, we fine-tune LLaVA-Next-Llama3-8B on our dataset and evaluate on MathVista~\cite{lu2023mathvista} (math images, testmini split), MMBench~\cite{liu2024mmbench} (mixed natural and scientific images, dev-en split), and RealworldQA~\cite{realworldqa} (natural images, test split). In Tab.~\ref{tab:common-vl-benchmark-evaluation}, our model leads to slight improvements on MathVista and RealworldQA and a decrease on MMBench. These early results might be inconclusive, and we plan to expand them in the future.

\begin{table}[ht]
\centering
\scriptsize
\setlength{\tabcolsep}{3.5pt}
\begin{tabular}{l|ccc}
\toprule
{Models} & {Mathvista~\cite{lu2023mathvista}} & {MMBench~\cite{liu2024mmbench}} & {RealWorldQA~\cite{realworldqa}} \\
\midrule
LLaVA-Next-Llama3-8B & 37.00 & \textbf{79.03} & 58.95 \\
+ ECD & \textbf{38.40} & 77.36 & \textbf{59.22} \\
\bottomrule
\end{tabular}
% }
\caption{GPT-Acc (\%) on 3 common vision-language benchmarks.}
\label{tab:common-vl-benchmark-evaluation}
\end{table}

\section{Analysis of the Generation Pipeline}
We construct several variants to demonstrate the effectiveness of the proposed pipeline.

\textbf{Separate data and code generation vs. joint generation} is shown in Tab.~\ref{tab:comparison-of-data-generation-strategy} (a) vs Tab.~\ref{tab:comparison-of-data-generation-strategy} (c). We separately generate the code and data, comparing it with the case where the code and data are generated together. We observe that ours is superior to generating data and code jointly, evidenced by the lower FID and higher entropy.

\textbf{Effectiveness of conditional and sequential data generation for multi-subplot charts.} We condition each subplot on previous ones to maintain consistency. We compare this method with joint multi-subplot generation in Tab.~\ref{tab:comparison-of-data-generation-strategy} (b) vs Tab.~\ref{tab:comparison-of-data-generation-strategy} (c). We observe a decline in the realism and complexity if we do joint generation, which indicates the effectiveness of our design. 

\begin{table}[]
\centering
\small
\begin{tabular}{l|cc}
\toprule
{Generation Strategy}       & {FID$\downarrow$}        & {Avg. Entropy$\uparrow$} \\ \hline
(a) Code and Data Joint Gen.                	&   {126.86 ± 3.11}   &   1.91           \\
(b) Subplots Parallel Gen. & {123.26 ± 2.15}          &    1.90          \\ \hline
(c) Ours & \textbf{120.63 ± 3.23} & \textbf{2.05}      \\ \bottomrule
\end{tabular}
\caption{Comparison of three chart generation strategies. (a) Generating chart code and data simultaneously. (b) Generate all the subplots data simultaneously. (c) Our data generation method (decoupling code and data; conditional generation of subplots). We define 12 different layouts 
% (i.e., (1, 2), (2, 1), (1, 3), (3, 1), (2, 
%  2), (1, 4), (2, 3),(2, 4), (4, 2), (3, 3), (3, 4), (4, 3) 
for the common ``bar+line'' combination, 
and generate 10 images for each layout. 
We use FID (with CharXiv)~\cite{salimans2016improved} and average pixel entropy to evaluate the realism and complexity of generated chart images. 
% We then calculate the Inception Score (IS)~\cite{salimans2016improved} to assess quality and evaluate image complexity and data coherence using the average GPT-based scoring [0-5].
}
\label{tab:comparison-of-data-generation-strategy}
\end{table}

\section{Numerical Diversity, Chart Style Statistics and Multi-Subfigure Reasoning Analysis} 
\textbf{Numerical Diversity.} Our current method has three strategies for numerical diversity. 1) We randomly sample data trend, \textit{e.g.}, increasing, decreasing, or stable. 2) we randomly sample the number of elements, \textit{e.g.}, the number of groups in a bar plot.  3) We use a high temperature (1.0) in GPT-4o to encourage varied parameter output.

\textbf{Chart Style Statistics.} We calculate the percentage of charts undergoing various changes using our synthesis method. For line-charts we find: 100.00\% charts have changes in line color, 99.65\% in font size, 95.09\% in line width, 67.37\% with non-default background, 59.30\% with transparency, 47.37\%  in grid, 39.65\% in new annotations, 29.12\% in area shading, 17.19\% include arrows, 11.23\% draw threshold lines, 10.18\% remove axis borders, 10.18\% add zoom-in insets and 3.51\% include error bars. In total, 58.95\% of line charts include 6 or more distinct changes.

\textbf{Multi-Subfigure Reasoning QA Pairs.} ~We analyze 500 reasoning QA pairs. $\sim$13.4\% are identified by GPT-4o as requiring multi-subfigure reasoning. Moreover, while some QAs can be answered by a single figure in a multi-subfigure chart, such QAs are still difficult because of the need for subfigure localization.

\section{Discussions and Limitations} 
\textbf{What makes a high-quality chart training set?} This paper mainly uses FID and average entropy as indicators to the similarity to real images and image complexity. However, the fact that ECD performs better than other datasets does not mean that ECD always has a lower FID and higher average entropy. We validated that the number of themes and chart types are also useful. Other factors such as text formats, font sizes, and even colors may also influence the training set quality. 

\textbf{Lack of a dedicated vision encoder for chart.} FID calculation requires feature extraction using InceptionNet-V3, pretrained on ImageNet (a natural image dataset). There are domain gaps between natural images and chart images, which might limit the efficacy of FID. The community still lacks a proper chart vision encoder to better compute indicators that require feature extraction.

\section{Comparison with Other Chart Datasets}
\begin{figure*}[t]
  \centering
   \includegraphics[width=0.93\linewidth]{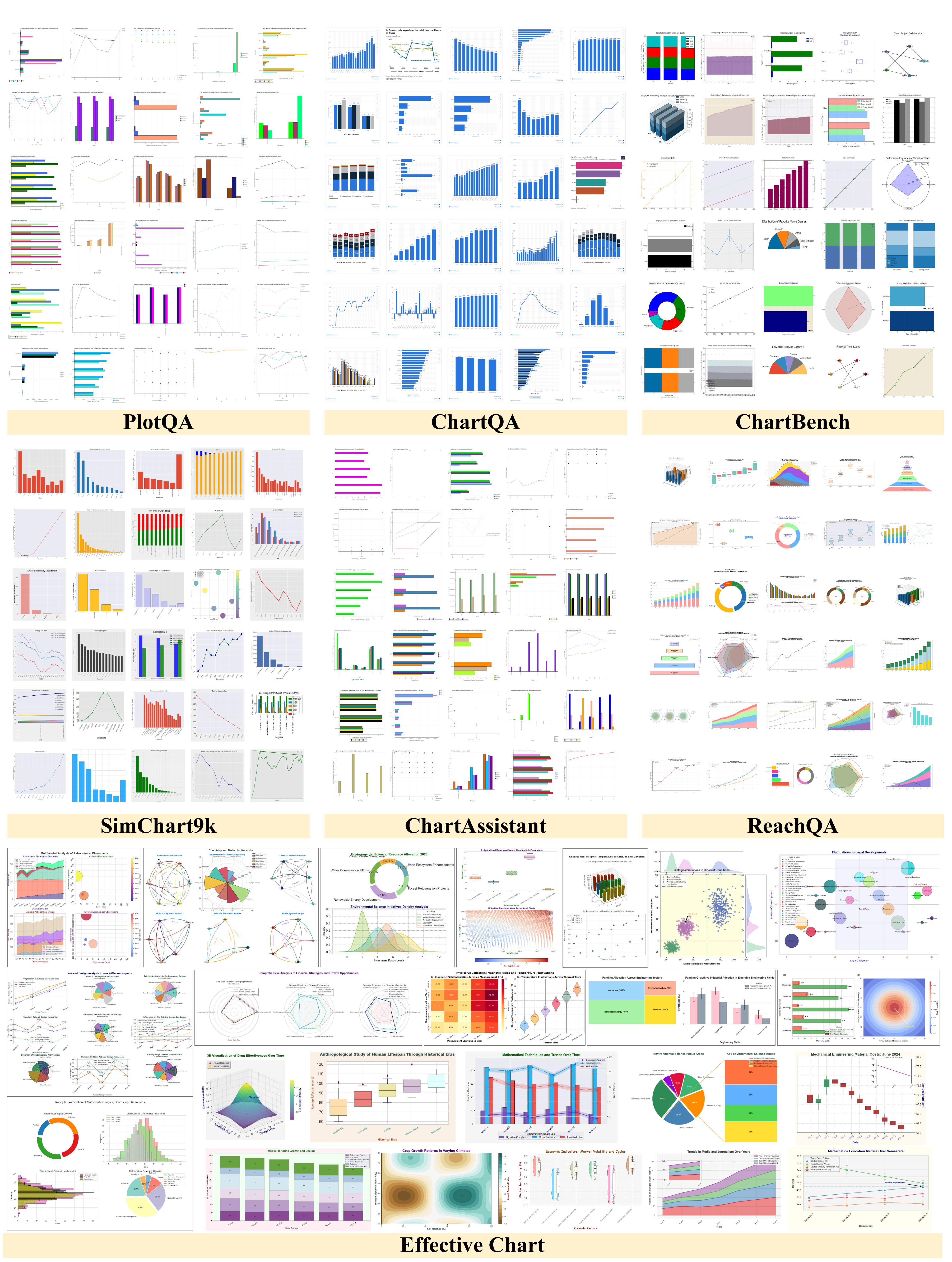}
   \caption{Compared to the chart images in existing chart QA training datasets (sampled for better visualization, \textit{i.e.}, PlotQA~\cite{methani2020plotqa} (3 types), ChartQA~\cite{masry2022chartqa} (3 types), ChartBench~\cite{xu2023chartbench} (9 types), SimChart9k~\cite{xia2023structchart} (3 types), ChartAssistant~\cite{meng2024chartassisstant} (9 types), ReachQA~\cite{he2024distill} (10 types)), it is evident that our ECD (29 types) exhibits a greater diversity of subplot combinations ($>$250 types of combination) and complexity.}
   \label{fig:visual-comparision-across-datasets.pdf}
\end{figure*}

To facilitate further comparison with other datasets, Fig.~\ref{fig:visual-comparision-across-datasets.pdf} present qualitative comparisons. This figure presents a visual comparison between our dataset's chart images and those from several chart QA training datasets, including PlotQA \cite{methani2020plotqa}, ChartQA \cite{masry2022chartqa}, ChartBench \cite{xu2023chartbench}, SimChart9k \cite{xia2023structchart}, ChartAssistant \cite{meng2024chartassisstant}, and ReachQA \cite{he2024distill}. Compared to these datasets, ours features a greater variety of chart types, more diverse combinations, more intricate details, and richer visual representations.

\section{Prompt for Chart Image Generation}
\begin{figure*}[t]
  \centering
   \includegraphics[width=1.0\linewidth]{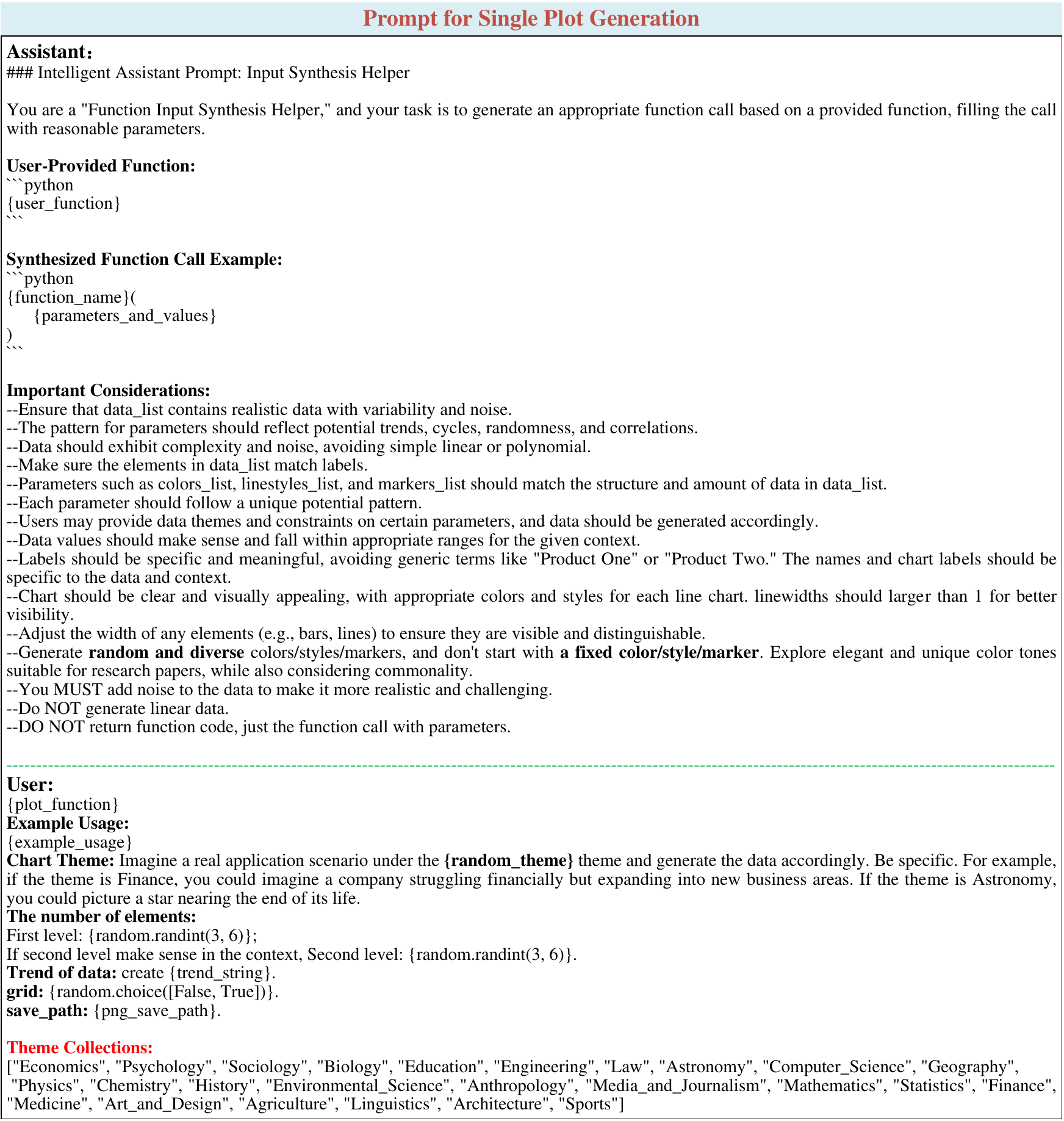}
   \caption{Prompt template for generating a single plot. We define 25 academic themes for generating charts across diverse disciplines.}
   \label{fig:prompt-for-single-plot-generation}
\end{figure*}

\begin{figure*}[t]
  \centering
   \includegraphics[width=1.0\linewidth]{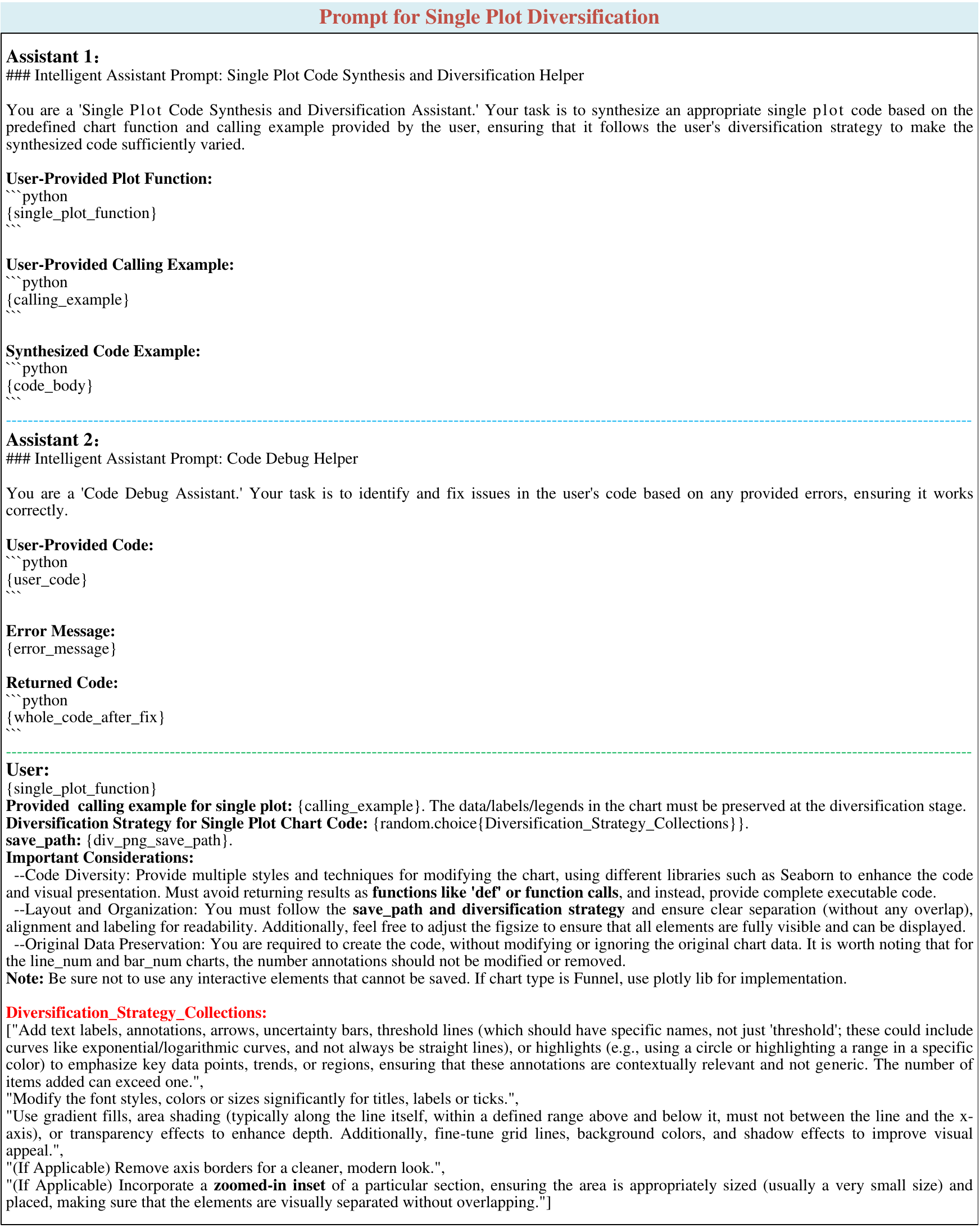}
   \caption{Prompt template for single plot diversification. We define 5 diversification strategies and randomly select one for each plot.}
   \label{fig:prompt-for-single-plot-diversification}
\end{figure*}

\begin{figure*}[t]
  \centering
   \includegraphics[width=1.0\linewidth]{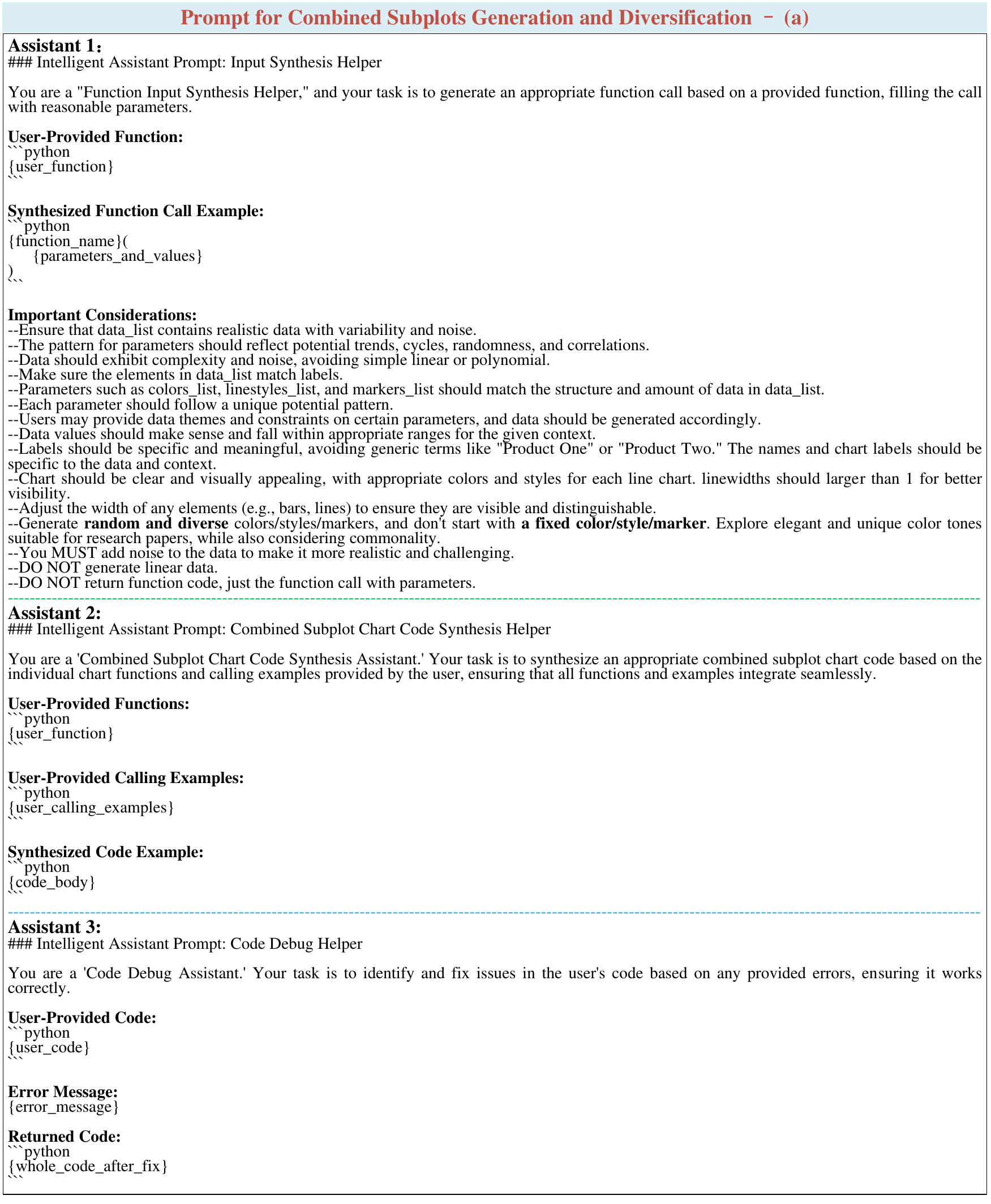}
   \caption{Prompt template for combined subplots generation and diversification - Part (a).}
   \label{fig:prompt-for-combined-subplots-generation-and-diversification-(a)}
\end{figure*}

\begin{figure*}[t]
  \centering
   \includegraphics[width=1.0\linewidth]{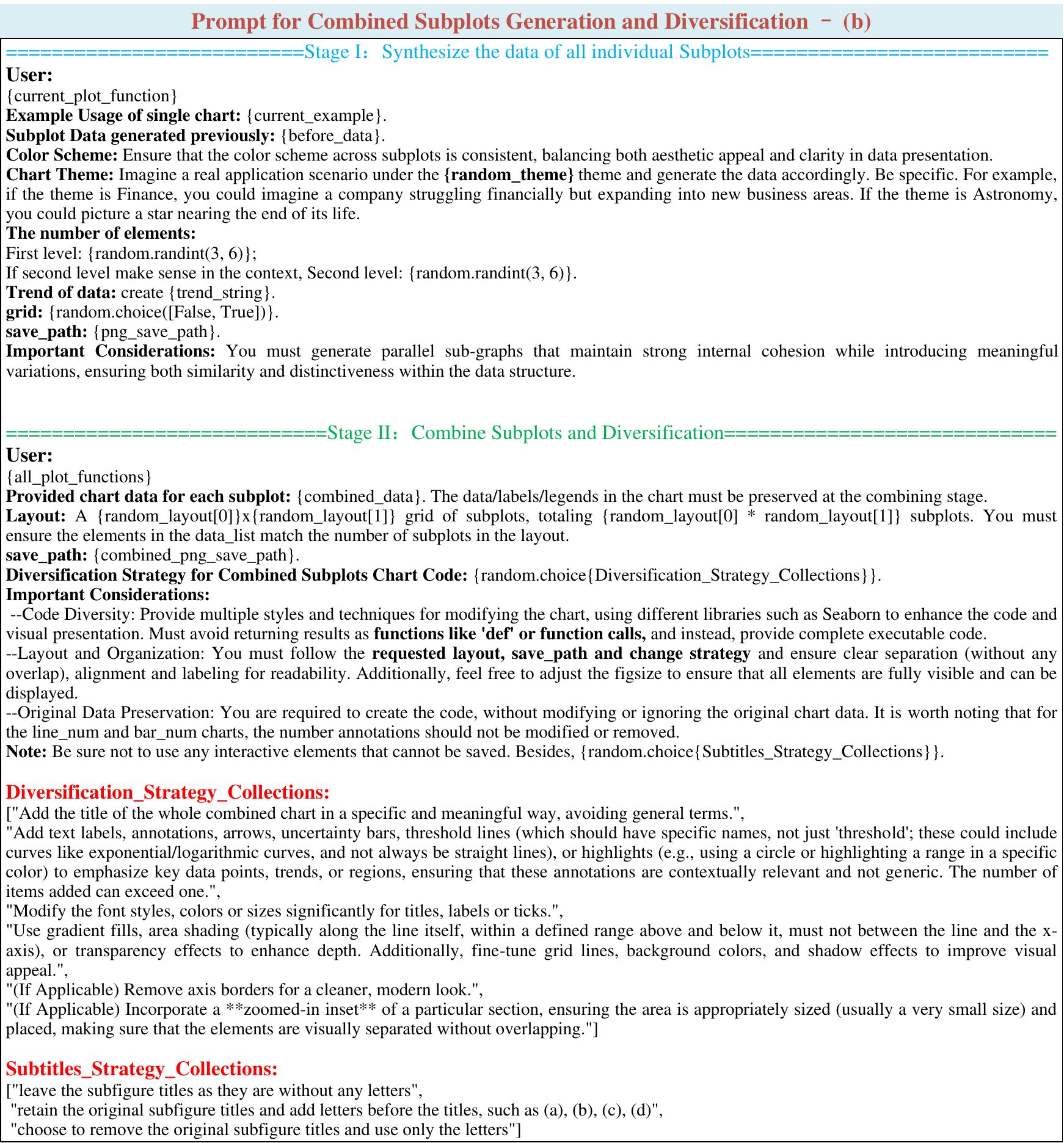}
   \caption{Prompt template for combined subplots generation and diversification - Part (b). The combined subplots introduce an additional diversification strategy on top of the single plot (i.e., add the title of the whole chart). Additionally, the diversification of subtitles is also randomly selected from the collection.}
   \label{fig:prompt-for-combined-subplots-generation-and-diversification-(b)}
\end{figure*}

\begin{figure*}[t]
  \centering
   \includegraphics[width=1.0\linewidth]{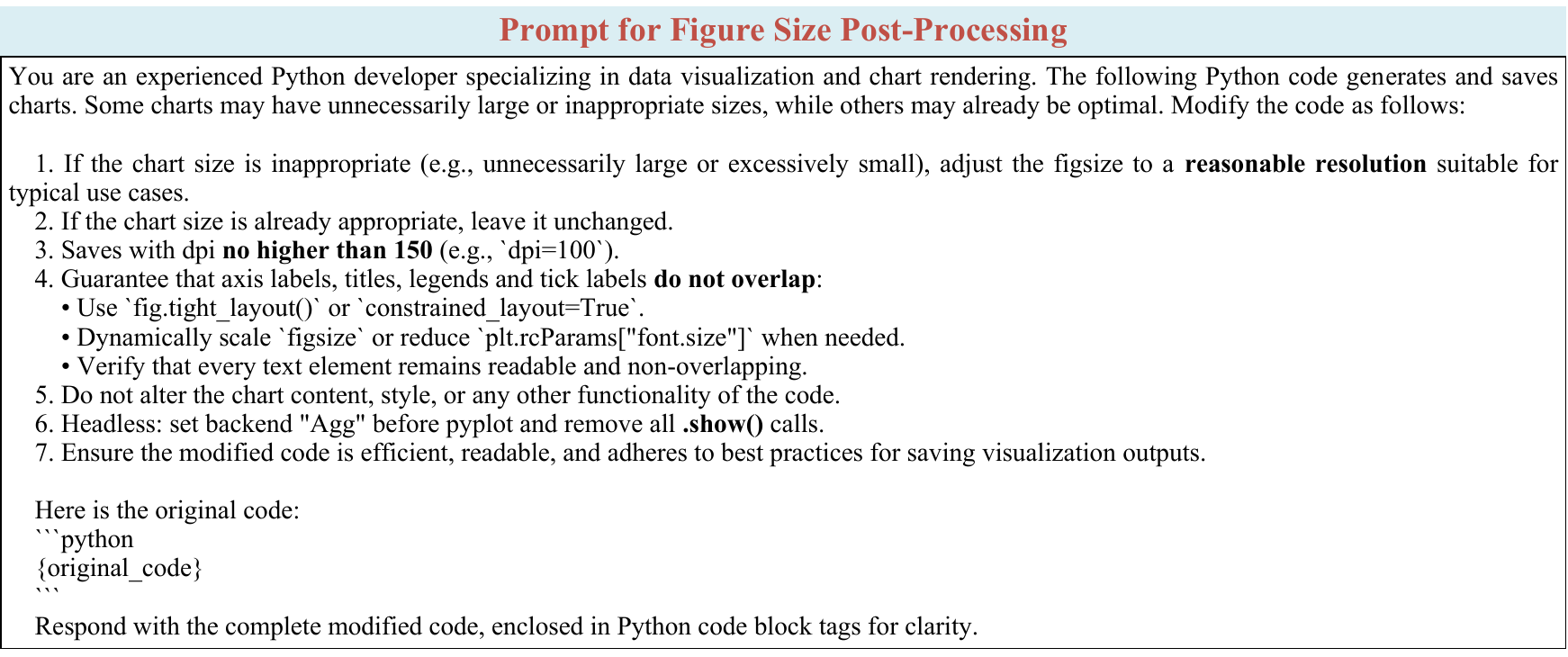}
   \caption{Prompt template for figure size post-processing. We enable GPT-4o to dynamically adjust figure size and visual elements during post-processing, ensuring better clarity and visualization.}
   \label{fig:prompt-for-figsize-post-processing}
\end{figure*}

Fig.~\ref{fig:prompt-for-single-plot-generation}, Fig.~\ref{fig:prompt-for-single-plot-diversification}, Fig.~\ref{fig:prompt-for-combined-subplots-generation-and-diversification-(a)} and Fig.~\ref{fig:prompt-for-combined-subplots-generation-and-diversification-(b)} illustrate the prompts we use to guide GPT-4o in generating single plot chart, combined subplot charts, and diverse visual variations. I) \textit{Single plot chart}: We employ predefined chart functions and examples to prompt GPT-4o to generate the data for a single plot chart. II) \textit{Combined subplot chart}: each single plot in the combined chart is generated sequentially based on conditional inputs (\textit{i.e.}, data of previous generated plots), ensuring semantic continuity across the entire chart composition. III) \textit{Diversification}: We define 5 and 6 diversification strategies for the single-plot chart and the combined subplot chart, respectively, and randomly select 1 strategy for each chart generation. It is important to note that additional prompts are incorporated to guide GPT-4o in prioritizing detail modifications while ensuring the preservation of the originally generated data. Fig.~\ref{fig:prompt-for-figsize-post-processing} presents the post-processing prompt applied to chart images, designed to empower GPT-4o in refining figure size and resolution. This step aims to enhance the conveyance of visual information, particularly in improving the clarity of elements such as text.

\section{Prompt for Chart Image Rating}
\begin{figure*}[t]
  \centering
   \includegraphics[width=1.0\linewidth]{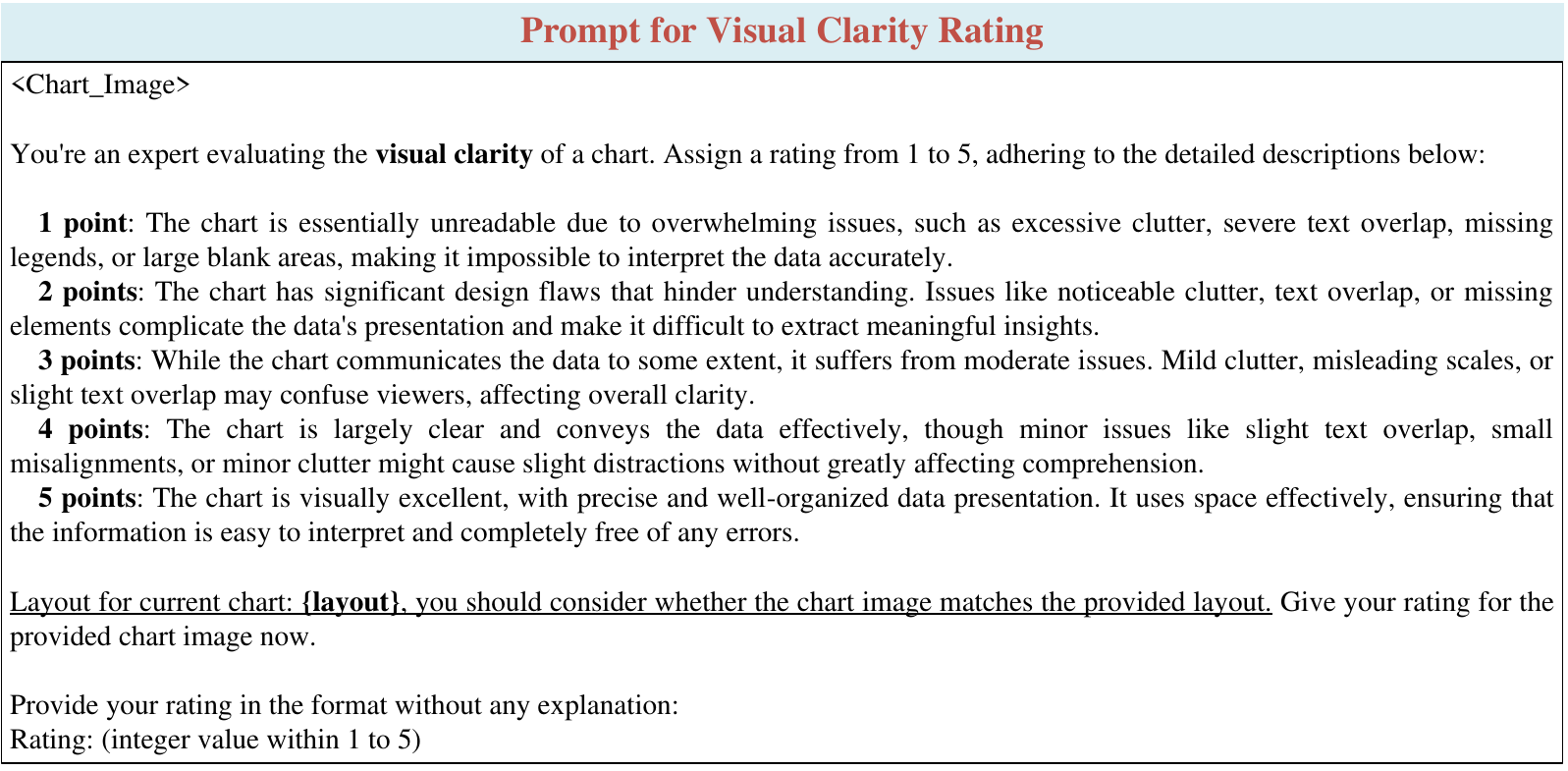}
   \caption{Prompt template for visual clarity rating. We apply a 1$\sim$5 point scale for GPT-4o to evaluate and select chart images with excellent visual clarity. The underlined portions are applicable only to combined subplot charts with layout that are not 1 by 1.}
   \label{fig:prompt-for-filtering-strategy-vc}
\end{figure*}

\begin{figure*}[t]
  \centering
   \includegraphics[width=1.0\linewidth]{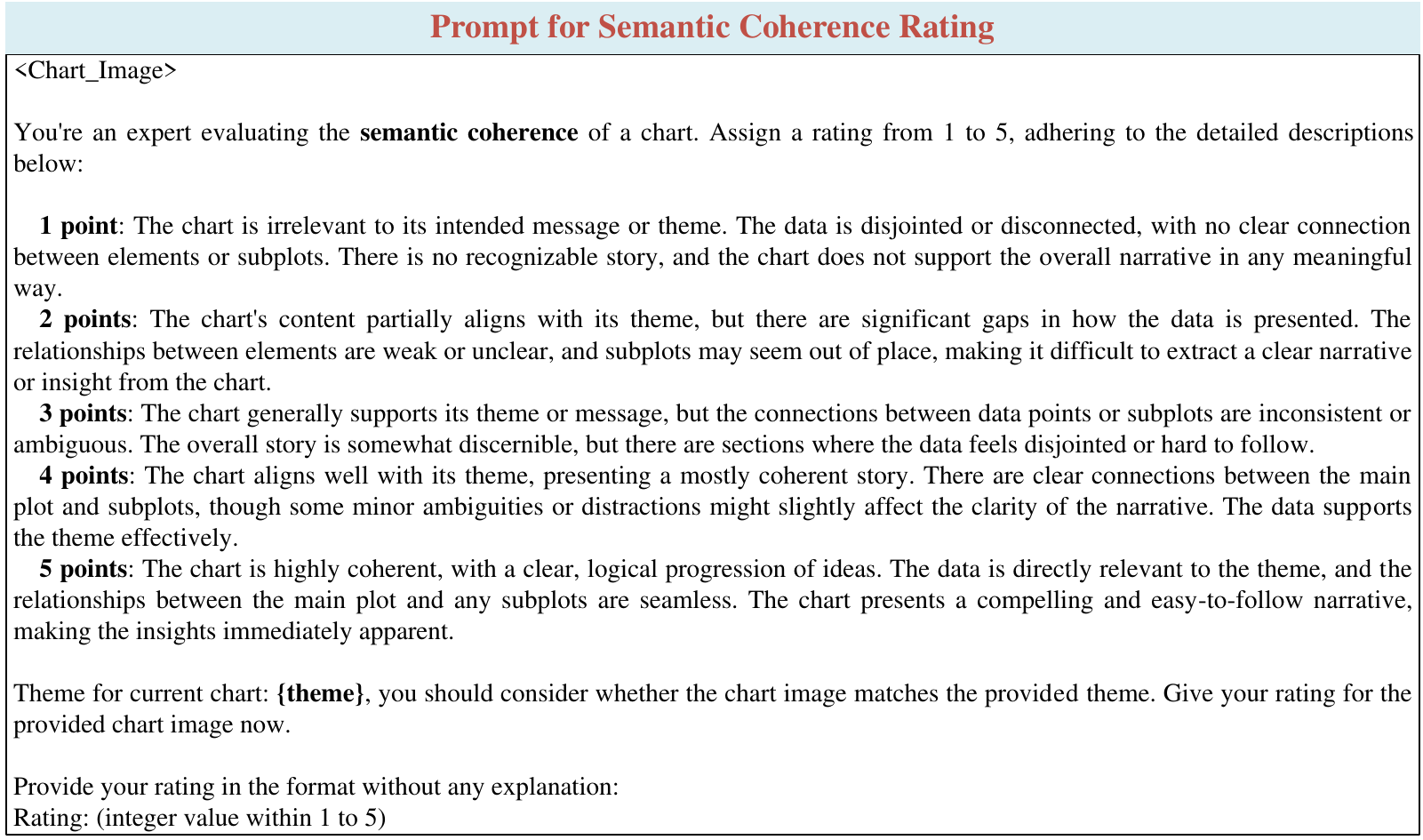}
   \caption{Prompt template for semantic coherence rating. We apply a 1$\sim$5 point scale for GPT-4o to evaluate and select chart images with clear, logical semantic coherence.}
   \label{fig:prompt-for-filtering-strategy-sc}
\end{figure*}

Fig.~\ref{fig:prompt-for-filtering-strategy-vc} and Fig.~\ref{fig:prompt-for-filtering-strategy-sc} illustrate the prompts utilized for chart image evaluation, incorporating two distinct rating criteria: ``visual clarity'' and ``semantic coherence''. The first criterion is employed to assess the visual integrity of the chart (on a scale from 1 to 5), specifically targeting the exclusion of images exhibiting excessive blank spaces or chaotic overlaps. The second criterion evaluates the semantic coherence (also on a scale of 1 to 5), ensuring that the chart adheres to a well-defined and logically structured theme. The integration of these two metrics serves to filter out charts of insufficient quality.

\section{Prompt for Chart QA Generation} 
\begin{figure*}[t]
  \centering
   \includegraphics[width=1.0\linewidth]{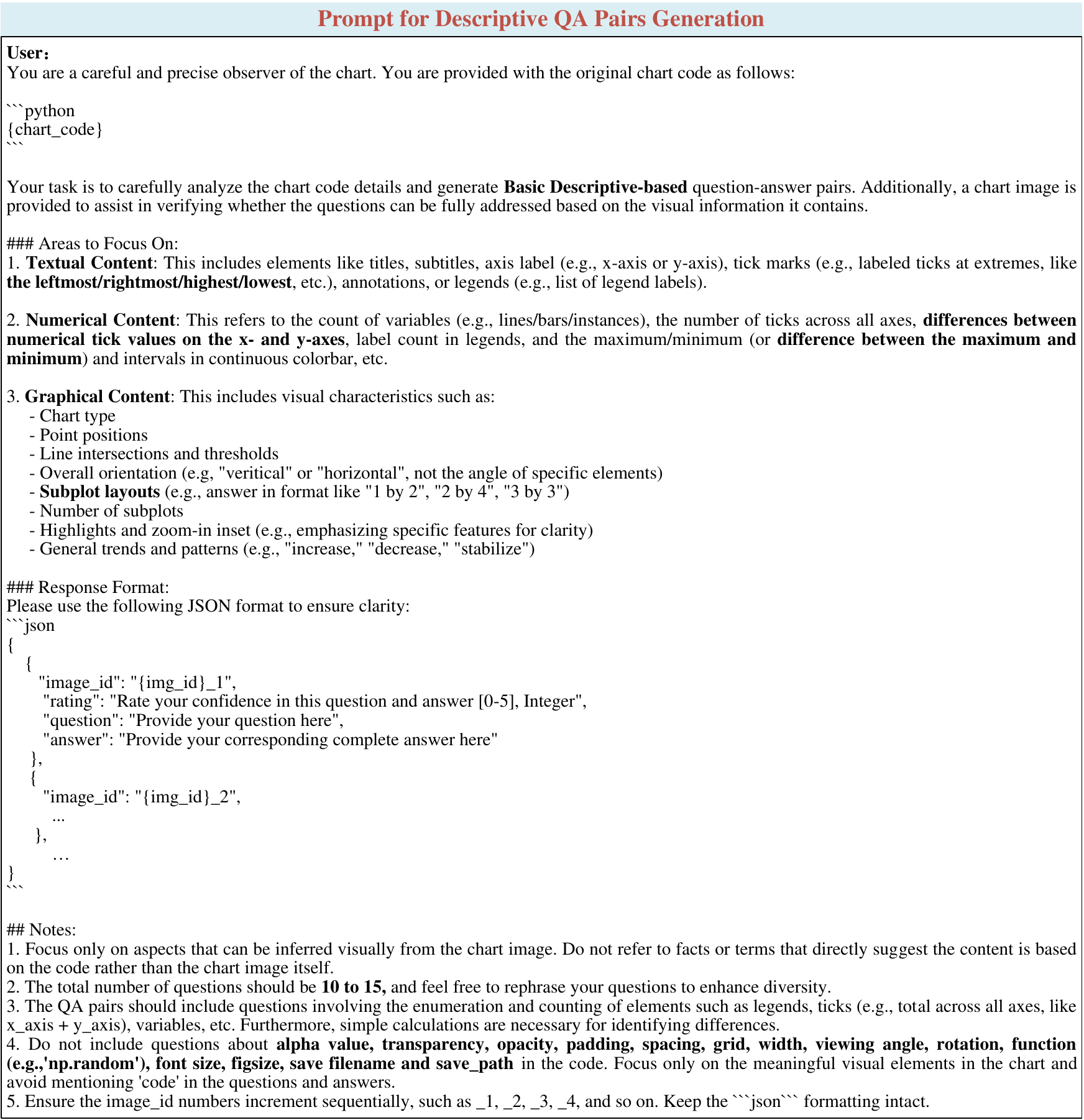}
   \caption{Prompt template for descriptive QA pairs generation. We prompt GPT-4o to focus on various aspects of descriptive queries, such as textual content, numerical content, and graphical content, and generate multiple question-answer pairs based on these dimensions.}
   \label{fig:prompt-for-descriptive-QAs}
\end{figure*}

\begin{figure*}[t]
  \centering
   \includegraphics[width=1.0\linewidth]{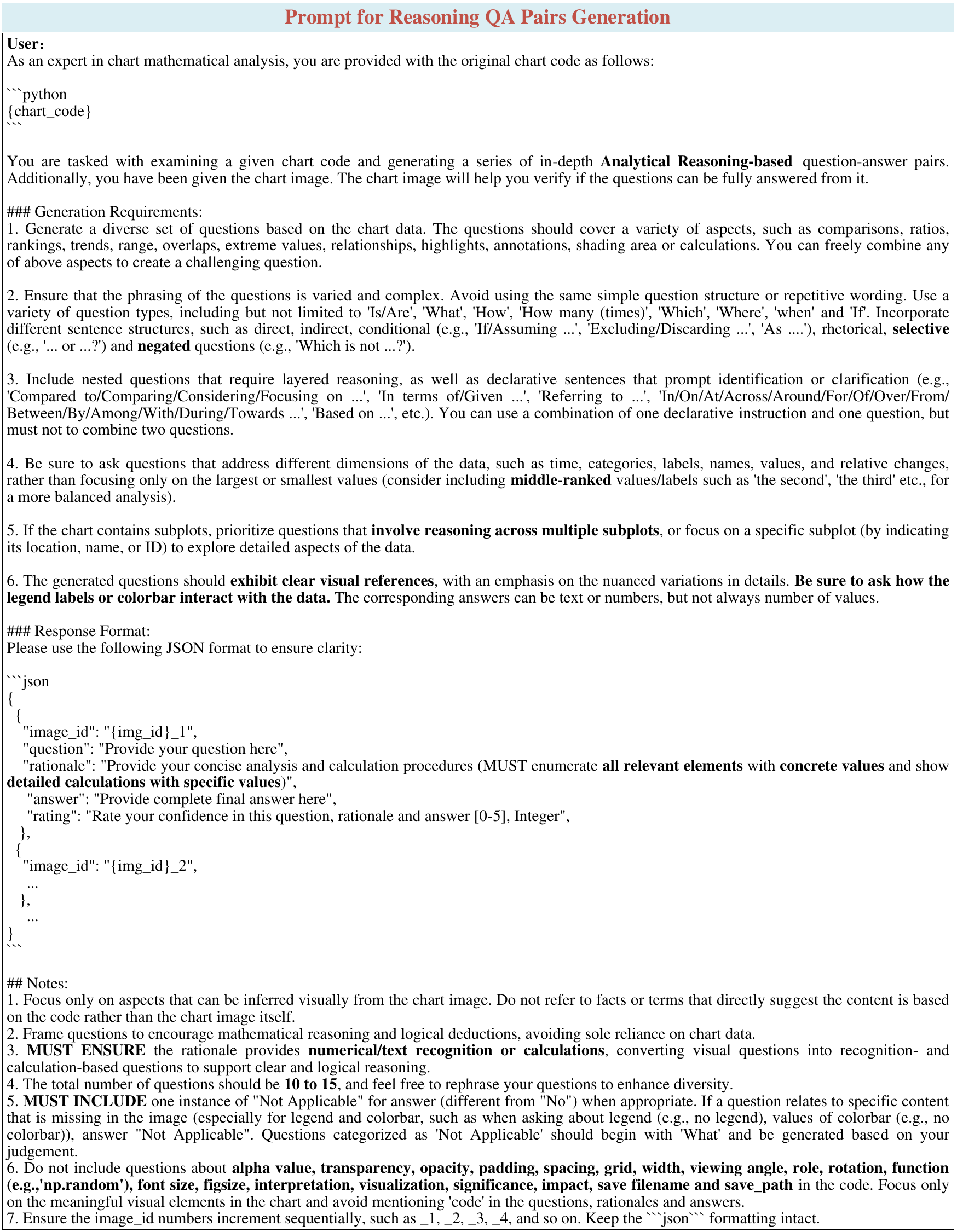}
   \caption{Prompt template for reasoning QA pairs generation. Prior to directly generating the final answer, we generate a rationale for in-depth analysis.}
   \label{fig:prompt-for-reasoning-QAs}
\end{figure*}

Fig.~\ref{fig:prompt-for-descriptive-QAs} and Fig.~\ref{fig:prompt-for-reasoning-QAs} illustrate the prompts we provide to GPT-4o for generating question-answer (QA) pairs based on chart code, where chart images are also incorporated as input to assist in determining whether the questions can be answered solely based on visual information. The generated QA pairs are categorized into two types: descriptive and reasoning. Descriptive QA pairs focus on the identification of fundamental textual, numerical, and graphical elements within the chart. Reasoning QA pairs, on the other hand, emphasize hierarchical analysis, computational inference, and logical deduction, with the generated answers incorporating rationale as part of the reasoning process. Additionally, for each generated QA pair, GPT-4o is tasked with producing a confidence rating, which serves as a measure for subsequent filtering.

\section{Prompt for Evaluation on ECDBench} 
\begin{figure*}[t]
  \centering
   \includegraphics[width=1.0\linewidth]{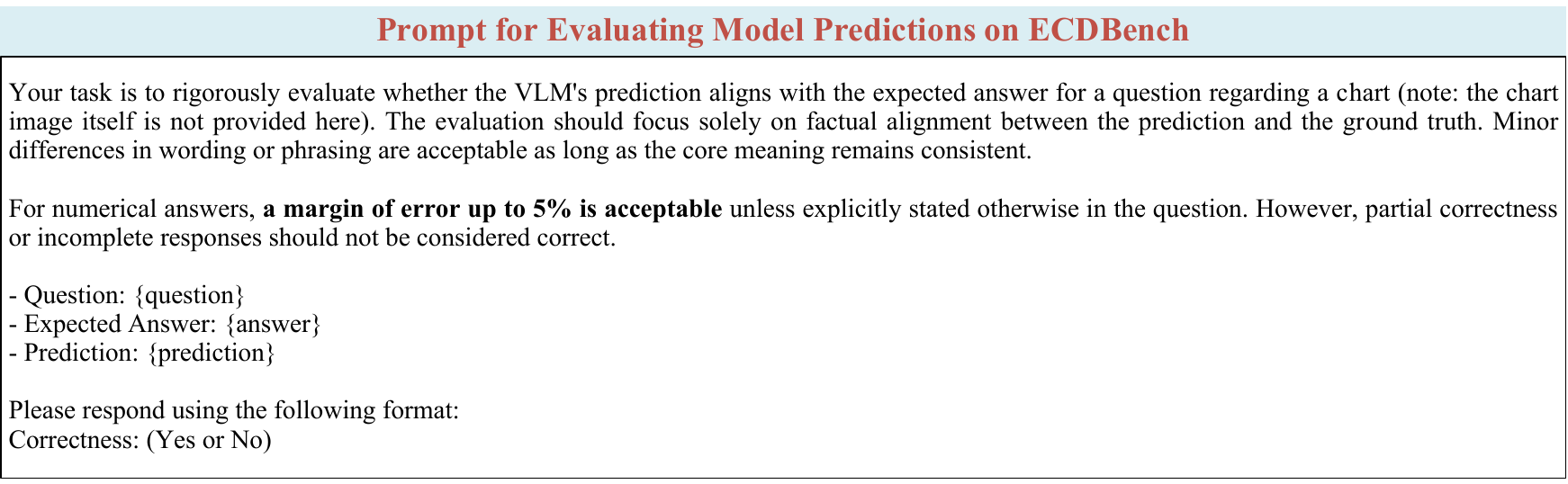}
   \caption{Prompt template for evaluating model predictions on our constructed ECDBench. For numerical responses, we allow a tolerance range of 5\% during evaluation.}
   \label{fig:prompt-for-ecdbench-evaluation}
\end{figure*}

Fig.~\ref{fig:prompt-for-ecdbench-evaluation} illustrates the prompt used for evaluating model predictions on ECDBench. Similarly, we employ the LLM-as-Judge approach to assess the responses. For answers containing numerical results, a 5\% tolerance is permitted to account for visual recognition discrepancies.

% \section{Rationale}

\end{document}